%% file: main.tex
\documentclass[nohyperref]{article}

\usepackage{microtype}
\usepackage{graphicx}
\usepackage{subfigure}
\usepackage{booktabs} %

\input{math_commands.tex}

\usepackage{hyperref}

\usepackage[accepted]{icml2022}

\usepackage{amsmath}
\usepackage{amssymb}
\usepackage{mathtools}
\usepackage{amsthm}
\usepackage{pifont}
\newcommand{\cmark}{\ding{51}}
\newcommand{\xmark}{\ding{55}}

\usepackage[capitalize,noabbrev]{cleveref}

\theoremstyle{plain}

\theoremstyle{definition}

\theoremstyle{remark}

\usepackage[textsize=tiny]{todonotes}

\icmltitlerunning{Perceiver AR}

\begin{document}

\twocolumn[
\icmltitle{General-purpose, long-context autoregressive modeling with Perceiver AR}

\icmlsetsymbol{equal}{*}
\icmlsetsymbol{equal2}{$\dagger$}

\begin{icmlauthorlist}
\icmlauthor{Curtis Hawthorne}{equal,google}
\icmlauthor{Andrew Jaegle}{equal,deepmind}
\icmlauthor{Cătălina Cangea}{deepmind}
\icmlauthor{Sebastian Borgeaud}{deepmind}
\icmlauthor{Charlie Nash}{deepmind}
\icmlauthor{Mateusz Malinowski}{deepmind}
\icmlauthor{Sander Dieleman}{deepmind}
\icmlauthor{Oriol Vinyals}{deepmind}
\icmlauthor{Matthew Botvinick}{deepmind}
\icmlauthor{Ian Simon}{google}
\icmlauthor{Hannah Sheahan}{deepmind}
\icmlauthor{Neil Zeghidour}{google}
\icmlauthor{Jean-Baptiste Alayrac}{equal2,deepmind}
\icmlauthor{Jo\~{a}o Carreira}{equal2,deepmind}
\icmlauthor{Jesse Engel}{equal2,google}
\end{icmlauthorlist}

\icmlaffiliation{deepmind}{DeepMind}
\icmlaffiliation{google}{Google Research, Brain Team}

\icmlcorrespondingauthor{Curtis Hawthorne}{fjord@google.com}
\icmlcorrespondingauthor{Andrew Jaegle}{\mbox{drewjaegle@deepmind.com}}

\icmlkeywords{Machine Learning, ICML, Perceiver AR, Perceiver, Language Modeling, Music Generation, Sequence Generation, ImageNet, Project Gutenberg, PG-19, Autoregressive Models, Generative Models}

\vskip 0.3in
]

\printAffiliationsAndNotice{\icmlEqualContribution} %

\begin{abstract}
Real-world data is high-dimensional: a book, image, or musical performance can easily contain hundreds of thousands of elements even after compression. However, the most commonly used autoregressive models, Transformers, are prohibitively expensive to scale to the number of inputs and layers needed to capture this long-range structure. We develop Perceiver AR, an \textbf{a}uto\textbf{r}egressive, modality-agnostic architecture which uses cross-attention to map long-range inputs to a small number of latents %
while also maintaining end-to-end causal masking. Perceiver AR can directly attend to over a hundred thousand tokens, enabling practical long-context density estimation without the need for hand-crafted sparsity patterns or memory mechanisms.
When trained on images or music, Perceiver AR generates outputs with clear long-term coherence and structure.
Our architecture also obtains state-of-the-art likelihood on long-sequence benchmarks, including 64~$\times$~64 ImageNet images and PG-19 books.
\end{abstract}

\section{Introduction}
A central goal of artificial intelligence research is the development of systems that can identify structure in the world and use it to effectively perform tasks of interest. In the past few years, autoregressive (AR) modeling with attention architectures (sometimes referred to metonymically as ``language modeling'') has emerged as a viable path to achieving this goal. In AR modeling, a set of outputs are generated by (i) using a model to map a set of inputs to an output, (ii) appending that output to the set of inputs, and proceeding again from step (i) until the full set of outputs has been produced. This simple recipe can in principle be followed to express any input-output relationship, and breakthrough results have been achieved using Transformers \cite{vaswani2017attention} or related models to learn the input~$\rightarrow$~output mapping \cite{vinyals2019grandmaster, brown2020language, jumper2021highly, wu2021nuwa}. For this to work, the model must be able to capture patterns in the input that are useful for predicting the next output. 

Patterns in real-world data often depend on the details of many inputs (or ``tokens,'' each of which represents a row of an input array), some of which are far away in space, time, or setting from the current output. Many pieces of music, for example, begin by stating a theme with clear melodic and rhythmic elements. Over the piece, these elements are gradually varied with increasingly elaborate restatements designed to draw listeners in. How does the precise timing of a phrase relate to its antecedents? How does a tonal motif persist and develop as it is restated? How does a new harmonization recontextualize a familiar melody? The structure of the piece emerges when each component is considered alongside many others.

There is a tension between this kind of long-form, contextual structure and the computational properties of Transformers. Transformers repeatedly apply a self-attention operation to their inputs: this leads to computational requirements that simultaneously grow quadratically with input length and linearly with model depth. As the input data grows longer, more input tokens are needed to observe it, and as the patterns in the input data grow more subtle and complicated, more depth is needed to model the patterns that result. Computational constraints force users of Transformers to either truncate the inputs to the model (preventing it from observing many kinds of long-range patterns) or restrict the depth of the model (denuding it of the expressive power needed to model complex patterns).

\textbf{The goal of this work} is to design an architecture that retains the well-known benefits of Transformers for autoregressive modeling while enabling long-range pattern recognition without adding extraneous complexity. To do this, we build on the  Perceiver family of attention architectures \cite{jaegle2021perceiver, jaegle2021io}, which have demonstrated excellent performance on a wide range of large-context domains. Perceivers use cross-attention to map a full input array into a smaller latent array and perform all subsequent attention operations in the resulting latent space. 
\begin{figure}[t]
    \centering
    \includegraphics[width=.85\columnwidth]{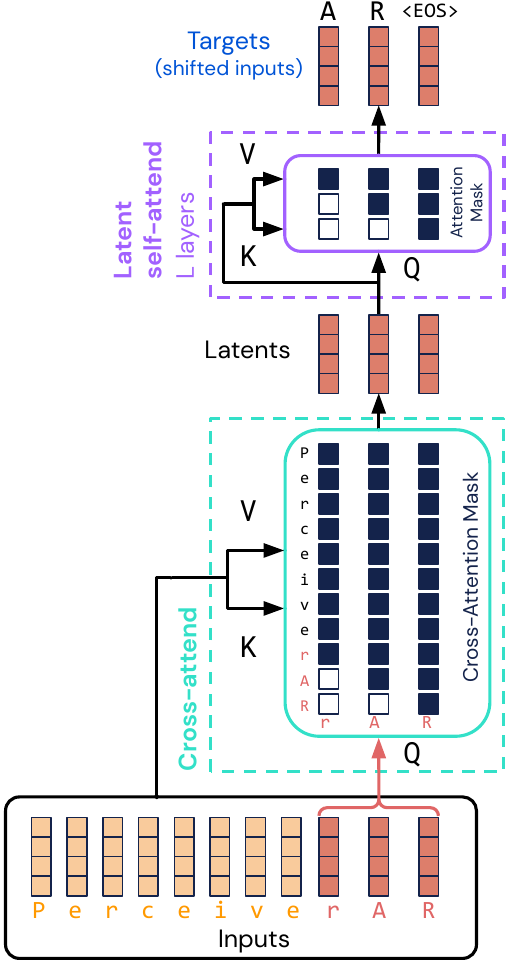}
    \caption{Perceiver AR maps inputs ($X \in \mathcal{R}^{M \times C}$; $M=11$ shown) to a small latent ($Z \in \mathcal{R}^{N \times C}$; $N=3$ shown) by cross-attention, querying with the $N$ most recent inputs to produce one latent for each target. Latents subsequently interact via a deep stack of $L$ self-attention layers to produce estimates for each target. Causal masking is used in both cross- and self-attention to maintain end-to-end autoregressive ordering.}
    \label{fig:perceiver_ar_architecture}
    \vspace{-15pt}
\end{figure}

This decouples the computational requirements of processing a large input array from those required to make a network very deep, allowing deep Perceivers to be used on large numbers of inputs. However, because each model latent attends to all inputs regardless of position, Perceivers cannot be used directly for autoregressive generation, which requires that each model output depend only on inputs that precede it in sequence.

Perceiver AR solves this problem with three fixes: (i) introducing an ordering to the latents by identifying each latent with a single output element, (ii) using causally masked cross-attention to allow each latent to attend only to input elements that precede it in sequence, and (iii) using causally masked self-attention throughout the latent processing stack to preserve this autoregressive dependency structure end-to-end. These changes allow Perceiver AR's outputs to be decoded in autoregressive sequence while preserving the essential computational and memory benefits of other Perceiver architectures. As each of the architecture's outputs is conditioned on all prior inputs, the architecture is well-positioned to capture long-range dependencies.

We show that this architecture produces excellent results on several real-world domains with long-range context: RGB-level images (\cref{sec:imagenet-64x64}), tokenized language (\cref{sec:pg19,sec:wikitext,sec:books}), and audio or symbolic music (\cref{sec:maestro}). Input lengths for these tasks are summarized in \cref{tab:model-context-length}. 
We demonstrate that Perceiver AR can learn to perfectly recognize long-context patterns over distances of at least 100k tokens on a synthetic copy task with known ground-truth structure (\cref{sec:copy_task_input_length}). We highlight several intriguing properties that result from decoupling input size from compute requirements: by keeping the long context, but changing the number of latents, we can (i) increase or decrease computational load at test time for improved (but slower) or faster (but somewhat worse) results (\cref{sec:compute_at_test_time}) or (ii) trade off model capacity against batch size at train time with no effect on test-time performance (\cref{sec:num_train_targets}).

We make the following contributions:
\begin{itemize}
    \item We introduce Perceiver AR, an efficient, domain-agnostic architecture for autoregressive generation that can directly attend to over a hundred thousand tokens.
    \item We demonstrate the utility of using long contexts for autoregressive generation: Perceiver AR obtains state-of-the-art results on ImageNet and Project Gutenberg density estimation and produces samples with a high degree of coherence and fidelity on several challenging generation tasks (images, symbolic music, and audio).
    \item We explore the benefits of decoupling the computational requirements of handling long inputs from those of model depth: improved efficiency compared to the widely used decoder-only Transformer and Transformer-XL architectures and the ability to vary the compute used at test time to match a target budget.
\end{itemize}

Model code is available at \url{https://github.com/google-research/perceiver-ar}.

\section{Autoregression and long-context modeling}

Autoregressive models (e.g. \citealt{schmidhuber1994sequential, rosenfeld2000two, bengio2003neural, graves2013generating, oord2016pixel, uria2016neural}) estimate the density of an example $X \in \mathcal{R}^{M}$ by decomposing it sequentially using the chain rule of probability:
\begin{align}
    p(X) = \prod_{m=0}^{M-1} p \Big( X_m \Big| X_{<m} \Big).
\end{align}
Each $X_m$ is typically a token (an audio sample, an RGB channel, a character, etc.). The chain rule tells us that the density of an input $X$ composed of arbitrarily many tokens can be estimated by sequentially estimating the conditional density of each of the $X$'s tokens. This requires that $p(X_m | \ldots)$ be conditioned on all prior tokens that are useful for predicting the $m$th token.%

Conditioning on all relevant inputs is challenging due to the difficulty of scaling standard models beyond small window lengths. In practice, models must be carefully designed so that the tokens that can be included in the context are as relevant as possible. This usually means that spatially or temporally local tokens are included in the context, but longer-range context is ignored or subsampled. As the context becomes smaller, this leads to worse approximations for signals that contain long-term dependencies.

Perceiver AR is designed to incorporate longer-term context into powerful density estimators, thereby allowing models to contextualize on more of a signal and giving better flexibility on how data is processed, moving us closer to the goal of general-purpose density modeling.

\begin{table}[t]
  \centering
  \begin{tabular}{c c}
    \toprule
    Task & Input Positions \\
    \midrule
    Copy (\cref{sec:copy_task}) & 131,072 \\
    ImageNet (\cref{sec:imagenet-64x64}) & 12,289 \\
    PG-19 (\cref{sec:pg19}) & 4,096 \\
    Books (\cref{sec:books}) & 16,384 \\
    Wikitext-103 (\cref{sec:wikitext}) & 8,192 \\
    Symbolic Music (\cref{sec:maestro}) & 65,536 \\
    Music Audio (\cref{sec:maestro}) & 65,536 \\
    \bottomrule
  \end{tabular}
  \caption{Maximum number of input (key-value) positions used for tasks in this paper. Most models used 1024 query positions.}
  \label{tab:model-context-length}
\end{table}

\section{Perceiver AR}
Perceiver AR follows Perceiver and Perceiver IO in addressing the problem of large inputs using a Transformer-style cross-attention module to map inputs $X \in \mathcal{R}^{M \times C}$ ($C$ is the number of channels) to a smaller number of latents $Z_1 \in \mathcal{R}^{N \times C}$:
\begin{align}
\label{eq:cross_attention}
    Z_1 \leftarrow \text{CrossAttend}(X, Z_0)
\end{align}
The latent array $Z_1$ is typically small ($N < M$), so it is amenable to processing by more self-attention modules: 
\begin{align}
\label{eq:cross_attention_2}
    Z_{l+1} \leftarrow \text{SelfAttend}(Z_l, Z_l).
\end{align}
This operation does not depend on the number of input points $M$, and $N$ can be chosen so that this operation is affordable and repeated for many layers $l \in [1, L]$. This strategy leads to an architecture (\cref{fig:perceiver_ar_architecture}) with complexity $\mathcal{O}(MN) + \mathcal{O}(LN^2)$ due to the cross-attention and the latent self-attention stack, respectively \cite{jaegle2021io}. For deep networks, the self-attention stack is where the bulk of compute occurs. %

\begin{figure}
    \centering
    \includegraphics[width=.9\columnwidth]{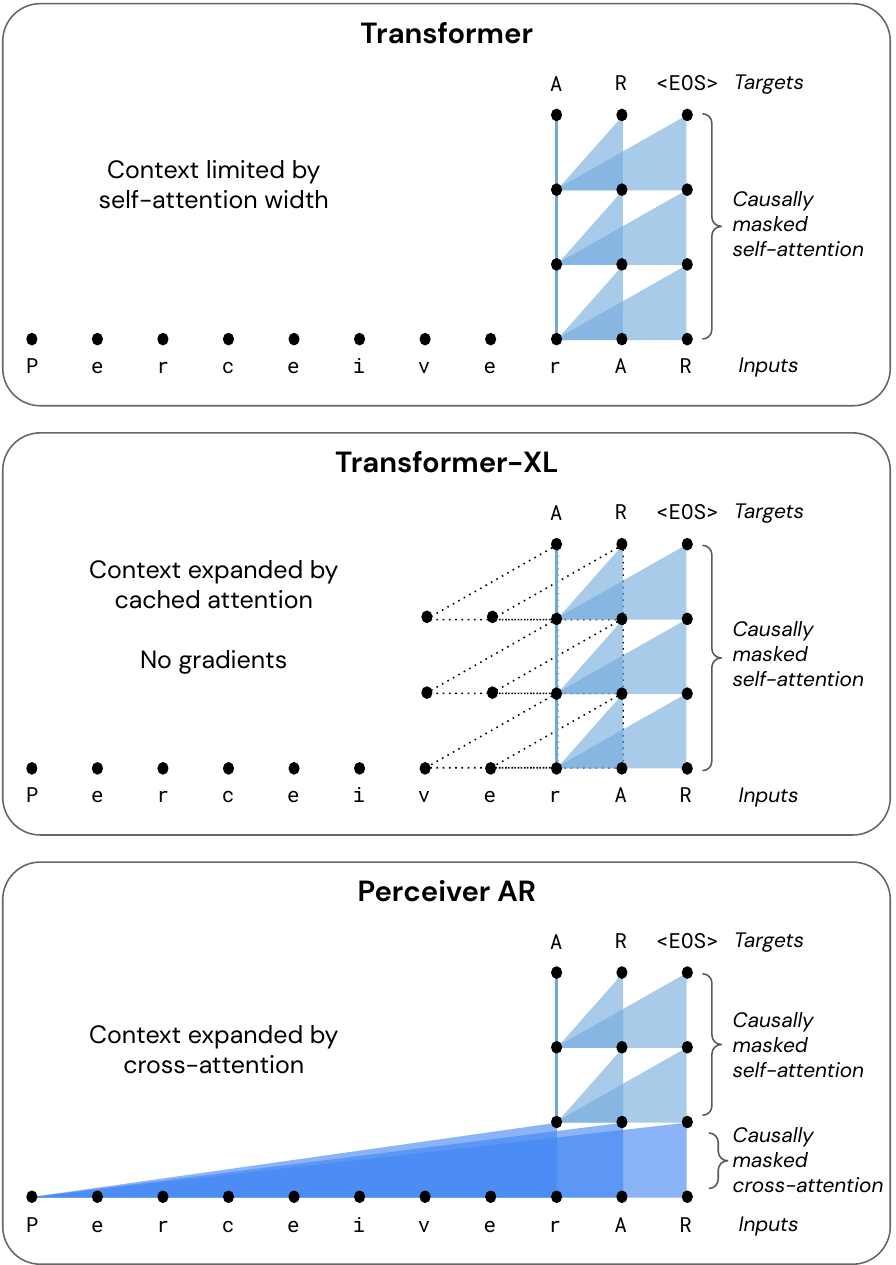}
    \caption{Perceiver AR compared to a standard decoder-only Transformer \cite{liu2018generating} and Transformer-XL (train-time configuration) \cite{dai2019transformerxl} during training. Only the subset of configurations that fit in device memory are shown. For a given self-attention stack width ($N=3$ shown here), the Transformer can process only the same number of input tokens. Transformer-XL incorporates more context while maintaining the width of the processing stack by caching and computing only the forward pass for longer-range inputs. In practice, Transformer-XL can incorporate only a moderate amount of additional context (see \cref{fig:perceiver_sps}). Perceiver AR uses a single masked cross-attend to enable training on much longer input contexts without requiring a wider self-attention stack.}
    \label{fig:architecture_comparison}
    \vspace{-15pt}
\end{figure}

But reducing the number of points from $M$ to $N$ prevents us from establishing the causal dependency between all input and output points used by Transformers for autoregressive modeling. Perceiver AR adapts the latents for autoregressive modeling by introducing causal masking to both the input cross-attention and to the latent self-attention layers (\cref{fig:perceiver_ar_architecture}) and assigning one latent to each of the final $N$ points of the input (i.e. those with the largest number of antecedents; $N=3$ in \cref{fig:perceiver_ar_architecture}). The influence of inputs that come after a given latent are masked at the cross-attention and all self-attention layers. 

To see how this works, consider the second (middle) latent in \cref{fig:perceiver_ar_architecture}: this latent is constructed by querying the input with \texttt{A}'s embedding. Because \texttt{R} follows \texttt{A} in sequence, \texttt{R}'s embedding is masked out, both at the cross-attention and at the subsequent self-attention layers. 

This procedure allows models to attend to long contexts while also preserving the autoregressive ordering of the targets through the entire network. The same procedure can be applied to any input that can be ordered, as long as masking is applied. For example, an image's RGB channels can be ordered in
raster scan order, by decoding the R, G, and B color channels for each pixel in the sequence (\cref{sec:imagenet-64x64}) or even under different permutations (\cref{sec:imagenet-rgb}).

\textbf{See~\cref{sec:perceiver_internals} for in-depth mathematical description of Perceivers and the Perceiver AR architecture and~\cref{sec:additional_details} for additional technical details.}

\section{Related work}
\subsection{Relationship to Transformer and Transformer-XL}

Perceiver AR decouples the length of the input from the computational requirements of the model, which are primarily controlled by the number of latents. 
In contrast, standard decoder-only Transformer architectures (\cref{fig:architecture_comparison}) maintain a one-to-one correspondence between inputs and outputs throughout the network, leading to $\mathcal{O}(LM^2)$ scaling. 

\begin{figure}[h]
    \centering
    \includegraphics[width=.9\columnwidth]{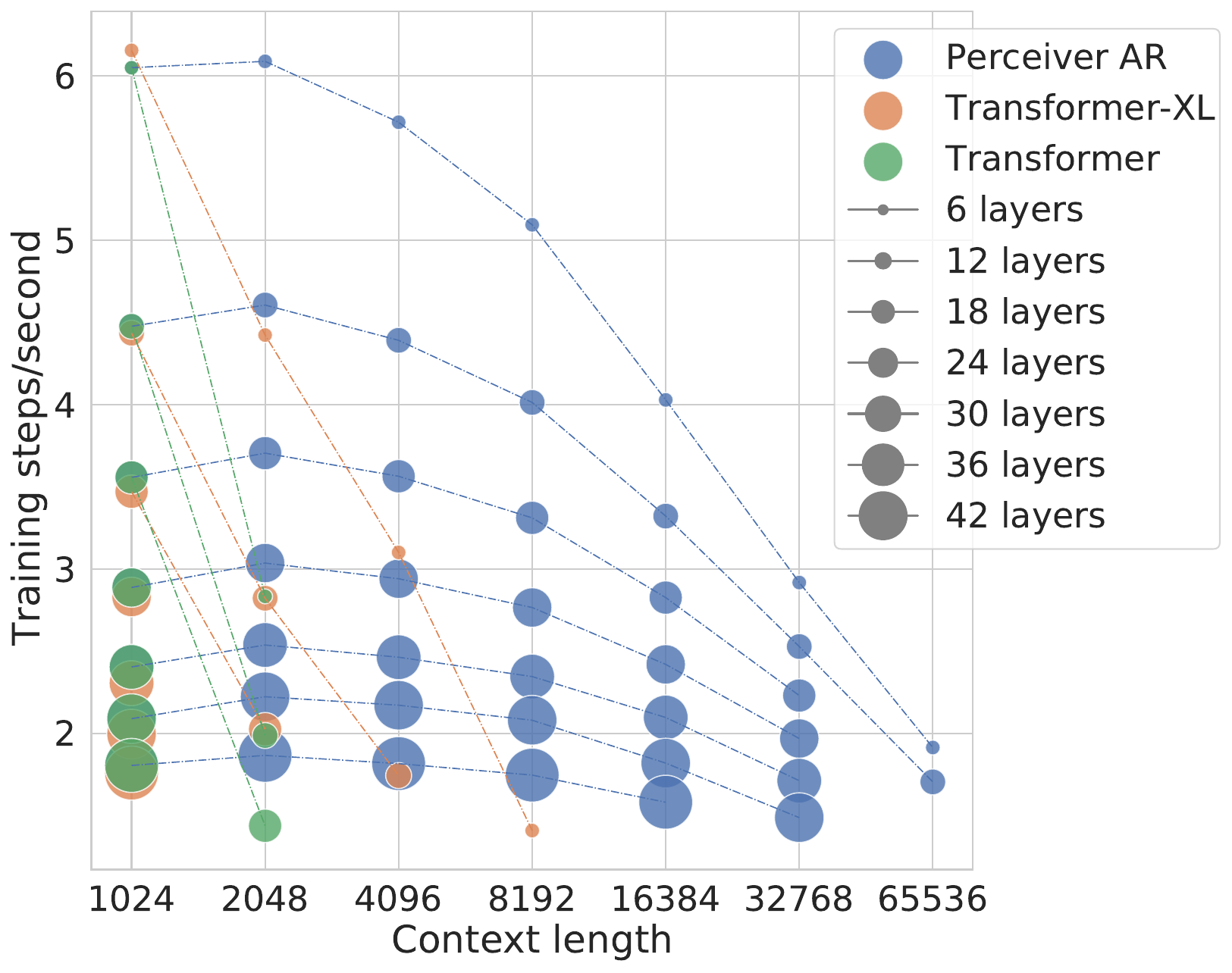}
    \caption{Training speed comparison on TPUv3 for Perceiver AR compared to a standard decoder-only Transformer and Transformer-XL. The Transformer is limited to a context length of 2,048 tokens, even with only 6 layers---larger models and larger context length require too much memory. Using the same 6-layer configuration, we can scale the Transformer-XL memory to a total context length of 8,192. Perceiver AR scales to 65k context length, and can be scaled to over 100k context with further optimization.}
    \label{fig:perceiver_sps}
    \vspace{-15pt}
\end{figure}

Perceiver AR also scales better to longer context in practice than Transformer-XL, perhaps the most commonly used method for extending context length in practice. Transformer-XL incorporates longer context by allowing attention over long-context positions at every layer in the forward pass and stopping gradient propagation to these positions in the backward pass. Transformer-XL also typically uses fewer positions at train than at test time, which further improves scaling at train time. But even with these modifications, the input size and model depth are still coupled, and as the context and depth increase, the forward pass becomes a compute and memory bottleneck. In practical terms, Perceiver AR scales better when both the context size and model depth increase. In  \autoref{fig:perceiver_sps}, we compare the wall-clock time per step of a decoder-only Transformer, Transformer-XL, and Perceiver AR in our codebase.

\subsection{Relationship to other scalable attention architectures}
Perceiver AR limits the computational requirements of processing long sequences by avoiding the use of one computational node for each input element. This strategy is also a feature of other architectures, like Set Transformer~\cite{lee2019set} and Luna~\cite{ma2021luna}, that use differently shaped query and key-value inputs to limit the cost of attention and produce ``downsampled'' intermediate representations that are smaller than the input's size. Unlike the Perceiver family of architectures, both Set Transformer and Luna alternate downsampling and upsampling attention operations throughout the architecture, which results in an architecture that mitigates some of the cost of standard Transformers, but still has compute requirements multiplicative in input size and model depth. Of this family of architectures, Luna is most similar, as it is compatible with (causally masked) autoregressive modeling. 

Several other architectures \citep{dai2020funnel, nawrot2021hierarchical, clark2021canine} reduce the processing requirements of Transformers by sequentially compressing the input with attention or convolution, but rely on the use of multiple large, quadratic-complexity attention layers or exploit locality assumptions that limit their generality. This makes them more efficient when applied to input chunks of similar size as normal Transformers (i.e. a few thousand), but reduces their utility for longer contexts.

\subsection{Relationship to encoder-decoder architectures}
Perceiver AR bears a resemblance to encoder-decoder Transformers \cite{vaswani2017attention}, seq2seq \cite{sutskever2014seq2seq}, and other encoder-decoder models. Encoder-decoder models pass long-context inputs into an encoder stack (e.g. \texttt{Perceive}, as in \cref{fig:perceiver_ar_architecture,fig:architecture_comparison}) and use the outputs of the encoder to contextualize the immediate antecedents of each output (e.g. \texttt{rAR}), which are processed by a separate decoder stack. In contrast, Perceiver AR passes both inputs and targets through a single, shared processing stack. This allows each target to learn how to use long-context and recent input as needed, with minimal architectural restrictions. From this point of view, Perceiver AR can be viewed as an encoder-decoder architecture with 0 encoder layers. By handling all inputs with a single (causally-masked) cross-attention, Perceiver AR sidesteps the need for separate encoder and decoder stacks.

\textbf{See \cref{sec:further_related_work} for an extended discussion of Perceiver AR's relationship to other methods for long-context modeling by efficient attention (\ref{sec:efficient_attention}), architecture design (\ref{sec:other_efficient_archs}), and input tokenization (\ref{sec:input_tokenization}).}

\section{Results}

We evaluate Perceiver AR on a number of different domains to demonstrate its flexibility and evaluate its ability to capture long-range structure on a variety of data. In all experiments except where mentioned, we use pre-layernorm attention modules \cite{xiong2020layernorm} and squared-ReLU nonlinearities \cite{so2021primer}. 

For each domain, we tuned models against eval perplexity with ad hoc hyperparameter sweeps as our compute permitted. We typically tuned channel size, head dimensions, and model depth. See each domain's section and appendices \ref{sec:training_details} and \ref{sec:eval_details} for more details.

\subsection{Copy Task}
\label{sec:copy_task}

\subsubsection{Long input length}
\label{sec:copy_task_input_length}

We first verify that the architecture can attend to very long input lengths on a synthetic copy task. Using a model with only 6 layers of 1024 latents, we provide an input with length $2^{17}$ (131,072). The model is trained on sequences containing a \texttt{[BOS]} (Beginning of Sequence) token followed by 65,535 random bytes (encoded as tokens taking one of 256 values). Those random bytes are then mirrored for the second half of the sequence and followed by an \texttt{[EOS]} (End of Sequence) token. This results in a maximum copy distance of $2^{17}-2$ tokens. Train and eval loss are calculated on only the second half of the sequence to avoid training on noise.

After training for 25k steps, the model was able to correctly predict the mirrored tokens plus \texttt{[EOS]} of 12 unseen validation sequences (a total of 786,432 tokens) with $100\%$ accuracy. This experiment demonstrates that the model can successfully attend to individual tokens within very long sequences and that a relatively small training signal of 1024 targets per sequence can successfully propagate through the cross-attend bottleneck even when most of the inputs are irrelevant for a given target. As an historical aside, we note that one antecedent of this experiment is the copy task used to validate the Neural Turing Machine \cite{graves2014neural} and Differentiable Neural Computer \cite{graves2016dnc}. These models showed nearly perfect accuracy when copying up to about length-50 sequences of random $2^{6}$- or  $2^{8}$-bit inputs, while Perceiver AR shows perfect accuracy when copying sequences of $2^{16}$ random $2^{8}$-bit tokenized inputs.

\subsubsection{Number of training targets}
\label{sec:num_train_targets}

In a standard decoder-only Transformer model, the input length, number of latent processing nodes per layer, and number of training targets are always the same. Perceiver AR allows the flexibility of any ratio of input length to number of latents and training targets. Changing the width of the self-attention stack affects the expressivity of the network and also alters the number of training outputs for which loss can be computed per sequence in a batch.

To illustrate this effect, we trained models on the copy task with a context length of 8,192 tokens using different numbers of latent nodes and batch sizes (\cref{tab:latents-targets}). To reduce the effect of network expressivity, we use only 1 self-attention layer for these experiments. We found that after training for 25k steps with 1024 latents and a batch size of 128, the model converged and predicted the second half of sequences in the test set with $100\%$ accuracy. If we reduced the batch size to 64, the model did not converge and achieved $<1\%$ accuracy on the test set. However, if we kept the batch size at 64 and increased the number of latents (and therefore training targets per sequence) to 2048, the model successfully converged.

\begin{table}[h]
  \centering
  \begin{tabular}{c c c}
    \toprule
    & 1024 latents & 2048 latents \\
    \midrule
    64 batch & \xmark & \cmark \\
    128 batch & \cmark & \cmark \\
    \bottomrule
  \end{tabular}
  \caption{Convergence on the copy task with a context length of 8,192 tokens. Either increasing the number of latents (and therefore number of targets) or batch size has a similar effect, as discussed in \cref{sec:num_train_targets}.}
  \label{tab:latents-targets}
\end{table}

\begin{table}[h]
  \centering
  \begin{tabular}{c c c}
    \toprule
    Model & Type & Bits/Dim \\
    \midrule
    PixelCNN & AR & 3.57 \\
    Sparse Transformer & AR & 3.44 \\
    Routing Transformer & AR & 3.43 \\
    Combiner & AR & 3.42 \\
    \textbf{VDM} & \textbf{Diff} & \textbf{3.40} \\
    \textbf{Perceiver AR (ours)} & \textbf{AR} & \textbf{3.40} \\
    \bottomrule
  \end{tabular}
  \caption{Results on Downsampled ImageNet (64~$\times$~64) density estimation in bits/dim on the validation set, lower is better. We compare our results against the autoregressive models PixelCNN \cite{oord2016pixel}, Sparse Transformer \cite{child2019generating}, Routing Transformer \cite{roy2021routing}, and Combiner \cite{ren2021combiner}, and the diffusion model Variational Diffusion Model \cite{kingma2021vdm}.}
  \label{tab:imagenet64-results}
\end{table}

Depending on memory, compute, and expressivity requirements, increasing the batch size or number of latents may be better: Perceiver AR enables this flexibility.

\subsection{ImageNet 64~$\times$~64}
\label{sec:imagenet-64x64}

\begin{figure*}[t]
    \centering
    \begin{tabular}{cc | cccc |cc}
    \includegraphics[width=45px]{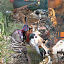} &
    \includegraphics[width=45px]{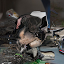} &
    \includegraphics[width=45px]{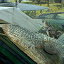} &
    \includegraphics[width=45px]{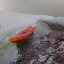} &
    \includegraphics[width=45px]{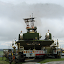} &
    \includegraphics[width=45px]{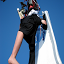} &
    \includegraphics[width=45px]{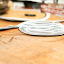} &
    \includegraphics[width=45px]{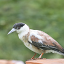}
    \\
    \includegraphics[width=45px]{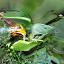} &
    \includegraphics[width=45px]{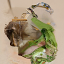} &
    \includegraphics[width=45px]{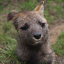} &
    \includegraphics[width=45px]{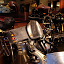} &
    \includegraphics[width=45px]{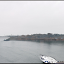} &
    \includegraphics[width=45px]{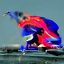} &
    \includegraphics[width=45px]{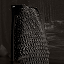} &
    \includegraphics[width=45px]{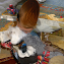}
    \\
    \multicolumn{2}{c}{16 latents} &
    \multicolumn{4}{c}{\textcolor{black}{1024} latents} &
    \multicolumn{2}{c}{1536 latents}
    \\
    \end{tabular}
    \caption{Representative samples from the ImageNet model. The 4 images on the left were generated using only 16 latents, the middle 8 with 1024 latents (same as \textcolor{black}{train time}), and the right 4 with 1536 latents. The same 60-layer model is used for all configurations and all can attend to the full sequence. The full batches from which these images were drawn are shown in \cref{sec:imagenet-samples}.}
    \label{fig:imagenet-samples}
    \vspace{-5pt}
\end{figure*}

To test this architecture's capabilities in the image modality, we use the downsampled ImageNet dataset \cite{oord2016pixel} at the 64~$\times$~64 resolution. Similar to the training procedure used in Sparse Transformer \cite{child2019generating}, we flatten the image into a sequence of color channel bytes (256 possible values) for each pixel in raster scan order. After adding a \texttt{[BOS]} token to the beginning of the sequence, this results in a length of 12,289. Each input has 3 randomly initialized position embeddings added to it for row (64), column (64), and color channel (3). No data augmentation is used.

\begin{table}[h]
  \centering
  \begin{tabular}{c c c}
    \toprule
    \# Latents & Bits/Dim & Generation (minutes) \\
    \midrule
    16 & 3.5664 & 1.99 \\
    64 & 3.4946 & 2.02 \\
    512 & 3.4139 & 2.81 \\
    1024 & 3.4025 & 3.68 \\
    1536 & 3.3986 & 4.69 \\
    2048 & 3.4018 & 5.88 \\
    4096 & 3.5569 & 12.28 \\
    \bottomrule
  \end{tabular}
  \caption{Results on Downsampled ImageNet (64~$\times$~64) density estimation in bits/dim on the validation set (lower is better) when changing the number of latents used at eval time (1024 used at \textcolor{black}{train time}). Generation time is how long a single image takes to infer on a TPUv3 core using activation caching described in \cref{sec:activation_caching_inference}. The same 60-layer model is used for all configurations and all can attend to the full sequence.}
  \label{tab:imagenet64-latents-results}
  \vspace{-10pt}
\end{table}

We train a model with 1024 latents in 60 self-attention layers after the initial cross-attend. After 750k steps, we achieve 3.40 bits/dim on the validation set, exceeding the performance of previous autoregressive models (\cref{tab:imagenet64-results}). Generated images samples are in \cref{fig:imagenet-samples} and \cref{sec:imagenet-samples}.

\subsubsection{Varying compute at test time}
\label{sec:compute_at_test_time}

Because Perceiver AR decouples input length from the number of latents in the self-attention stack and because no position-specific parameters are learned in this model, we have the intriguing option of evaluating with a different number of latents than were used during training, while still maintaining the ability to attend to the full input sequence (illustrated further in \cref{fig:test_time_configs}).

\begin{table}[h]
  \centering
  \begin{tabular}{c c c}
    \toprule
    Input Context & Standard Ordering & $R\to G\to B$  \\
    \midrule
    1024 &  3.55 & 4.63\\
    12289 & 3.54 & 3.53\\
    \bottomrule
  \end{tabular}
  \caption{Impact of context length and sequence ordering for ImageNet (64~$\times$~64). Results are bits/dim on the validation set, lower is better. ImageNet examples contain 12,288 ($64 \times 64 \times 3$) tokens. For short contexts, the sequence ordering has a large effect on performance. For long contexts, the effect is small.}
  \label{tab:imagenet-rgb-results}
  \vspace{-10pt}
\end{table}

\begin{table*}[t]
  \centering
  \begin{tabular}{c c c c c}
    \toprule
    Model & Context length & \# layers & Val ppl. & Test ppl. \\
    \midrule
    Transformer-XL \cite{rae2019compressive} & 512+1024 & 36 & 45.5 & 36.3 \\
    Compressive Transformer \cite{rae2019compressive} & 512+512+2x512 & 36 & 43.4 & 33.6  \\
    Routing Transformer \cite{roy2021routing} & 8192 & 22 & - & 33.2 \\
    \midrule
    Perceiver AR (ours) & 2048 & 60 & 45.9 & \textbf{28.9} \\
    Perceiver AR (ours) & 4096 & 60 & 45.9 & 29.0 \\
    \bottomrule
  \end{tabular}
  \caption{Results on PG-19 language modeling. Results are shown in perplexity (ppl.), lower is better. Baseline results are reproduced from the original papers. Routing Transformer does not report validation set performance. The context lengths shown include memory (Transformer-XL and Compressive Transformer) and compressive memory (the latter only).} 
  \label{tab:pg_19_results}
  \vspace{-12pt}
\end{table*}

We found that increasing the number of latents up to 2x the number used during training improves model performance, and decreasing the number latents results in gracefully degrading performance despite dramatic reductions in compute requirements, as seen in \cref{tab:imagenet64-latents-results}. For example, when using only 16 latents to attend to the full input sequence of 12,289 tokens, the model achieves the same bits/dim as PixelCNN \cite{oord2016pixel}. This kind of flexibility enables a single model to be used in scenarios with varying compute, latency, and quality requirements, without requiring additional training or a distillation process.

We suspect these models could be made even more flexible to the number of latents used during evaluation or inference if variable latent access is incorporated into training, but we leave a full exploration of these ideas to future work.

\subsubsection{Impact of sequence ordering and length}
\label{sec:imagenet-rgb}

We test Perceiver AR's ability to utilize context beyond the width of its self-attention stack by training on image sequences with strong long-range dependencies. Typically autoregressive models of image data use a raster scan ordering, where RGB subpixels in each image location are generated in sequence before moving on to the next location. We re-order ImageNet image data so that all the red subpixels are predicted in sequence, then the green, then the blue. This induces strong long range dependencies between subpixels in the same spatial location. 

We train 16-layer versions of the models in \cref{sec:imagenet-64x64} for 30k steps (due to compute constraints), and compare short context models (1024 inputs) to full context models (12,289 inputs). As a baseline we also train both model variants on sequences with the standard ordering. The results are shown in Table \ref{tab:imagenet-rgb-results}. We find that for the standard ordering, both small and long context models perform similarly. However, on the re-ordered image data, the short context model has significantly worse performance. Whereas the long context model has comparable performance to the standard ordering baselines. This indicates that our long context model is able to access and process information at distant timesteps. 

For general domains, we often do not know which long-range dependencies are important, and this experiment shows that the performance impact of missing existing dependencies can be severe. Perceiver AR provides a solution by extending Transformer context size in a scalable way.     

\subsection{Project Gutenberg (PG-19)}
\label{sec:pg19}

We next test the architecture for language modeling on the Project Gutenberg (PG-19) dataset \cite{rae2019compressive}, a collection of English-language books published before 1919 and scraped from the \href{https://www.gutenberg.org/}{Project Gutenberg eBook depository}. We use PG-19 as it is publicly available and contains a large number of words (1.97B train, 3.01M validation, 6.97M test) drawn from a reasonably large number of books (28k train, 50 validation, 10 test). We evaluate Perceiver AR on PG-19 using Subword-tokenized inputs as in prior work \cite{rae2019compressive, roy2021routing}. We use the Subword tokenization settings reported in section 4.2 of \citet{rae2019compressive}, and we trained models until the loss converged on a subset of the validation set (after training on about 200k steps at batch size 2048, or about 420B total tokens).

We compare Perceiver AR to state-of-the-art numbers from the literature (\cref{tab:pg_19_results}). The best models reported by these papers use 36 (Compressive Transformer; \citealt{rae2019compressive}) and 22 layers (Routing Transformer; \citealt{roy2021routing}). We find that Perceiver AR outperforms state-of-the-art methods on this dataset when using the same tokenization. However, consistent with previous work, we saw no evidence of improvement beyond 2k tokens on PG-19 \cite{sun2021long_range}.

Motivated by early experiments, both models use large input embeddings (4096), large numbers of cross-attend heads (128), and high cross-attend dropout (0.96875 for the 4096-context model and 0.875 for the 2048-context model; see \cref{sec:cross_attend_dropout}). The results presented in \cref{tab:pg_19_results} suggest that simply scaling Perceiver AR can produce excellent results even with very high levels of cross-attend dropout.

Consistent with prior work, we notice that models exhibit a consistent drop in performance between PG-19 validation and test sets.
This is likely due to the relatively small number of books contained in the PG-19 validation (50 books) and test (100 books) sets. Despite the large number of tokens, the small number of books on the validation set will lead many of these tokens to exhibit shared content and style, limiting how representative the validation set can be of the task as a whole. As noted by \citet{sun2021long_range}, the PG-19 validation set contains at least one book (out of 50) that is arguably out of distribution for the train set. 

To better understand the effect of context in language modeling, we next evaluate on a large internal book dataset where the number of documents is not a concern (\cref{sec:books}).

\subsection{Books}
\label{sec:books}

Here, we study the usefulness of longer input context by training on an internal dataset containing 4 million books published between 1500 and 2008. This dataset was previously used in~\citealt{rae2021scaling} as part of the MassiveTest dataset. We train models with 1024 latents, 36 layers and $\{1024, 4096, 8192, 16384\}$ input context tokens. %

These results (\cref{tab:books_results}) show a clear trend: better results are obtained with contexts longer than 1024.

\begin{table}[H]
  \centering
  \begin{tabular}{c c c c}
    \toprule
    Model & Context & Eval ppl. & Train Steps/sec \\
    \midrule
    Perceiver AR & 1024 & 14.88 & 2.19 \\ %
    Perceiver AR & 4096 & 14.60 & 2.09 \\ %
    Perceiver AR & 8192 & 14.57 & 1.95 \\ %
    Perceiver AR & 16384 & 14.56 & 1.75 \\ %
    \bottomrule
  \end{tabular}
  \caption{Books results, shown in perplexity (ppl.), lower is better.}
  \label{tab:books_results}
  \vspace{-15pt}
\end{table}

We also run experiments where we compare our model against a Transformer-XL baseline, with compute (measured by steps per second) matched as closely as possible. Models are trained with a batch size of 256 for 500k steps.

First, we evaluate 36-layer Perceiver AR models with $\{1024, 4096, 8192, 16384\}$ context lengths, respectively. Each of them is matched with a Transformer-XL trained on sequences of length 1024 and $\{23, 24, 25, 28\}$ layers, respectively. Table~\ref{tab:books_ar_36l_matched_by_txl_per_context} shows that our model improves consistently over the Transformer-XL in this controlled scenario.

\begin{table}
  \centering
  \begin{tabular}{c c c |c c}
    \toprule
    Model & Context & Depth & Steps/Sec & Eval ppl. \\
    \midrule
    AR & 1024 & 36 & 2.19 & 14.006 \\
    T-XL & 1024 & 23 & 2.17 & 14.822 \\
    \midrule
    AR & 4096 & 36 & 2.09 & 13.806 \\
    T-XL & 1024 & 24 & 2.06 & 14.719 \\
    \midrule
    AR & 8192 & 36 & 1.95 & 13.791 \\
    T-XL & 1024 & 25 & 1.97 & 14.593 \\
    \midrule
    AR & 16384 & 36 & 1.75 & 13.749 \\
    T-XL & 1024 & 28 & 1.76 & 14.276 \\
    \bottomrule
  \end{tabular}
  \caption{Books results on the test set, shown in perplexity (ppl.), lower is better, from 36-layer Perceiver AR models with varying contexts and Transformer-XL models with depth matched for compute (steps/sec), trained for 500k steps.}
  \label{tab:books_ar_36l_matched_by_txl_per_context}
\end{table}

\begin{table}
  \centering
  \begin{tabular}{c c c |c c}
    \toprule
    Model & Context & Depth & Steps/Sec & Eval ppl. \\
    \midrule
    T-XL & 1024 & 42-layer & 1.17 & 13.253 \\
    \midrule
    AR & 1024 & 62-layer & 1.19 & 12.849 \\
    AR & 4096 & 61-layer & 1.21 & 12.680 \\
    AR & 8192 & 60-layer & 1.25 & 12.660 \\
    AR & 16384 & 56-layer & 1.26 & 12.816 \\
    \bottomrule
  \end{tabular}
  \caption{Books results on the test set, shown in perplexity (ppl.), lower is better, from a 42-layer Transformer-XL model and Perceiver AR models with varying contexts and depth matched for compute (steps/sec), trained for 500k steps.}
  \vspace{-15pt}
  \label{tab:books_42l_txl_matched_by_ar_per_context}
\end{table}

In Table~\ref{tab:books_42l_txl_matched_by_ar_per_context}, we then compare the deepest Transformer-XL model (42 layers) that fits in memory against 4 Perceiver AR models with the same context lengths and respective numbers of self-attention layers $\{62, 61, 60, 56\}$. Remarkably, we were not able to increase the number of layers for some AR models---and achieve a closer match in compute---without running out of memory. Instead, we ran the deepest possible models for a specific context length. Even in this scenario, all Perceiver AR models performed better than the deepest Transformer-XL.

\subsection{Wikitext-103}
\label{sec:wikitext}
We further evaluate on the Wikitext-103 \citep{merity2016pointer} dataset, a commonly used word-level language modeling benchmark. The dataset consists of 28,475 Wikipedia articles containing between 68 and 26,993 words, averaging at 3.6k words. Wikitext-103 is a small dataset where strong regularization is required to prevent severe overfitting and to obtain good performance. Nonetheless, we obtain competitive results, suggesting Perceiver AR's utility even for relatively small datasets.

\begin{table}[h]
  \centering
  \begin{tabular}{c c c}
    \toprule
    Model & Valid ppl. & Test ppl. \\
    \midrule
    Adaptive inputs & 18.0 & 18.7 \\
    Transformer-XL & 18.3 & 18.2 \\
    Shortformer & 17.5 & 18.15 \\
    Compressive Transformer & 16.0 & 17.1  \\
    Routing Transformer & - & 15.8 \\
    \midrule
    Transformer-XL (ours) & 17.58 & 18.42 \\
    Perceiver AR (1024) & 17.86 &  18.52 \\
    Perceiver AR (2048) & 17.60 & 18.35 \\
    Perceiver AR (4096) & 17.66 & 18.25 \\
    Perceiver AR (8192) & 17.58 & 18.37 \\
    \bottomrule
  \end{tabular}
  \caption{Results on the Wikitext-103 language modeling benchmark. Baseline results are reproduced from the original papers: Adaptive inputs \cite{baevski2018adaptive},  Transformer-XL \cite{dai2019transformerxl}, Shortformer \cite{press2021shortformer}, Compressive Transformer \cite{rae2019compressive}, and Routing Transformer \cite{roy2021routing}. Transformer-XL (ours) is a reimplementation in our codebase. We train Perceiver AR models with varying context lengths.}  
  \label{tab:wikitext103}
  \vspace{-8pt}
\end{table}

We show results in \cref{tab:wikitext103}. Perceiver AR performs on par with Transformer-XL Large \citep{dai2019transformerxl} and Shortformer \cite{press2021shortformer}. We also include the state of the art results on Wikitext-103 (trained on Wikitext-103 only) \citep{rae2019compressive, roy2021routing}. Increasing the context length from 1,024 to 2,048 for Perceiver AR does improve perplexity but further increasing to 4,096 or 8,192 is harmful. This effect has been noted in previous work \citep{press2021shortformer, sun2021long_range} and further provides evidence that current language model performance on small benchmarks is not bottlenecked by their limited range context.

\subsection{MAESTRO}
\label{sec:maestro}

\begin{table}
  \centering
  \begin{tabular}{cccc}
    \toprule
    Model & MAESTRO & Test & Validation \\
    \midrule
    Music Transformer & v1 & - & 1.84  \\
    Perceiver AR & v1 & 1.82 & 1.82  \\
    \hline
    Perceiver AR & v3 & 1.91 & 1.90  \\
    \bottomrule
  \end{tabular}
  \caption{Results on MAESTRO symbolic music generation on Test and Validation datasets. Results are shown in negative log-likelihood, lower is better. Baseline result is reproduced from~\cite{hawthorne2018enabling}.}
  \label{tab:maestro_symbolic_results}
  \vspace{-10pt}
\end{table}

We also evaluate Perceiver AR in the music domain, with samples available in the online supplement (\url{https://bit.ly/3uF5LJg}). Here, modeling long-range dependencies is essential for obtaining good representations \cite{huang2018music}. We use the MAESTRO dataset~\cite{hawthorne2018enabling}, which contains approximately 200 hours of music in both symbolic (MIDI) and audio modalities.
For symbolic data, we experiment with both v1 and v3 versions of MAESTRO. The former allows us to compare with the existing state-of-the-art; the latter is an improved set with approximately 10\% more performances. We use the MIDI tokenization described in Section A.2 of~\citet{huang2018music}.

After training for 1M steps with an input context of 4096 tokens and 2048 latents in 12 self-attention layers, our model obtains a lower negative log-likelihood (\cref{tab:maestro_symbolic_results}) than Music Transformer~\cite{huang2018music, hawthorne2018enabling}. That model had 6 layers and was trained on random crops of 2048 tokens, but benefited from data augmentation, which ours did not. We also report results on MAESTRO v3.

For audio data, we generate vector-quantized embeddings using the SoundStream neural audio codec~\cite{zeghidour2021soundstream} at several bitrates.
Lower-bitrate codecs model coarser structure and enable training on a longer time window for a fixed context length, but at the expense of lower audio fidelity. We show results in~\cref{tab:maestro_v3_ss_results}.%

\subsection{Music samples} %
\label{sec:music_samples}

To showcase the long-term coherence of these models, we introduce another symbolic task involving a larger, private dataset of piano performances
\cite{simon2019}.
These were transcribed from 10,000+ hours of audio, using a variation of the Onsets and Frames model~\cite{hawthorne2017onsets}. From this dataset, we only used pieces resulting in 1024--32,768 tokens. Samples shorter than the lower limit are unlikely to contain song content; at the other end, there are only about 200 pieces longer than 32,768 tokens. The same tokenization as for MAESTRO is used. We train a model with 1024 latents and 24 self-attention layers on input contexts of 32,768 tokens, achieving a negative log-likelihood of 1.24 on the test set.

\begin{table}
  \centering
  \begin{tabular}{c c c c}
    \toprule
    SoundStream bitrate & Context & Test & Validation \\
    \midrule
    12kbps & 54.4s & 2.49 & 2.34  \\
    18kbps & 36.8s & 2.60 & 2.57  \\
    22kbps & 29.6s & 2.65 & 2.62  \\
    \bottomrule
  \end{tabular}
  \caption{Perceiver AR negative log-likelihood results on SoundStream audio generation, for a fixed context length of 65536.} \label{tab:maestro_v3_ss_results}
  \vspace{-20pt}
\end{table}

We generate samples from the models trained on this new task, as well samples from MAESTRO v3 symbolic and SoundStream models. The samples obtained from the large transcription dataset exhibit stylistic and structural coherence that spans several minutes, containing repeating musical themes, chord patterns, arpeggios and even ritardandos.
The audio domain samples exhibit the same tradeoff of audio fidelity and long-term structure seen in the previous section, but demonstrate strong correspondence to the original domain.
Rendered audio samples are available in the online supplement:
\url{https://bit.ly/3uF5LJg}.

\section{Conclusion}
We introduce Perceiver AR, an architecture designed for long-context autoregressive modeling. Perceiver AR scales to longer input sizes than the architectures typically used in practice (Transformers and Transformer-XL), while scaling to the depths needed for density estimation on real-world data. Perceiver AR decouples the computational requirements of processing many inputs from those of building deep networks. This gives us more control over how much compute is used for a given model at test time and allows us to smoothly trade off speed against performance. Perceiver AR produces good results on a number of domains. Our experiments suggest that the larger context used by Perceiver AR may open the door for more flexible autoregressive ordering strategies. Perceiver AR is a good candidate for a general-purpose, long-context autoregressive model.

\section{Acknowledgements}
Thanks to Chitwan Saharia for help with SR3 upsampling, Christos Kaplanis for Books guidance, and Jordan Hoffmann and Richard Tanburn for help with RoPE. Thanks to Daniel Toyama, Douglas Eck, Adam Roberts, Nando de Freitas, Dani Yogatama, Josh Gardner, Catalin Ionescu, Skanda Koppula, Rabeeh Karimi Mahabadi, Ethan Manilow, Viorica Patraucean, Oleh Rybkin, Nikolay Savinov, and others at DeepMind and Google Research for suggestions.

\bibliography{bibliography}
\bibliographystyle{icml2022}

\clearpage
\newpage
\appendix

\section{ImageNet Samples}
\label{sec:imagenet-samples}

\begin{figure*}[t]
    \centering
    \begin{tabular}{cccccccc}
    \includegraphics[width=45px]{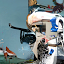} &
    \includegraphics[width=45px]{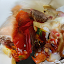} &
    \includegraphics[width=45px]{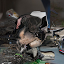} &
    \includegraphics[width=45px]{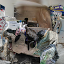} &
    \includegraphics[width=45px]{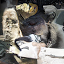} &
    \includegraphics[width=45px]{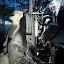} &
    \includegraphics[width=45px]{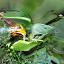} &
    \includegraphics[width=45px]{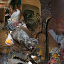}
    \\
    \includegraphics[width=45px]{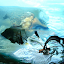} &
    \includegraphics[width=45px]{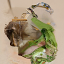} &
    \includegraphics[width=45px]{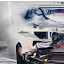} &
    \includegraphics[width=45px]{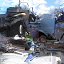} &
    \includegraphics[width=45px]{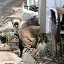} &
    \includegraphics[width=45px]{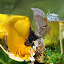} &
    \includegraphics[width=45px]{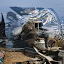} &
    \includegraphics[width=45px]{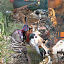}
    \\
    \end{tabular}
    \caption{Full batch of generated samples from the model trained on ImageNet using \textbf{16 latents} during inference.}
    \label{fig:imagenet-samples-z16}
\end{figure*}

\begin{figure*}[t]
    \centering
    \begin{tabular}{cccccccc}
    \includegraphics[width=45px]{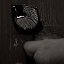} &
    \includegraphics[width=45px]{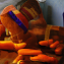} &
    \includegraphics[width=45px]{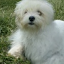} &
    \includegraphics[width=45px]{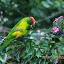} &
    \includegraphics[width=45px]{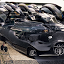} &
    \includegraphics[width=45px]{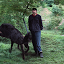} &
    \includegraphics[width=45px]{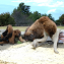} &
    \includegraphics[width=45px]{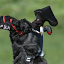}
    \\
    \includegraphics[width=45px]{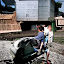} &
    \includegraphics[width=45px]{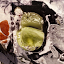} &
    \includegraphics[width=45px]{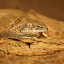} &
    \includegraphics[width=45px]{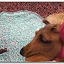} &
    \includegraphics[width=45px]{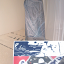} &
    \includegraphics[width=45px]{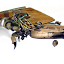} &
    \includegraphics[width=45px]{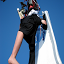} &
    \includegraphics[width=45px]{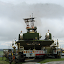}
    \\
    \includegraphics[width=45px]{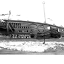} &
    \includegraphics[width=45px]{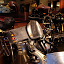} &
    \includegraphics[width=45px]{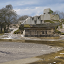} &
    \includegraphics[width=45px]{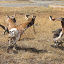} &
    \includegraphics[width=45px]{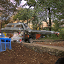} &
    \includegraphics[width=45px]{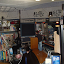} &
    \includegraphics[width=45px]{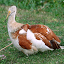} &
    \includegraphics[width=45px]{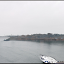}
    \\
    \includegraphics[width=45px]{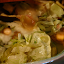} &
    \includegraphics[width=45px]{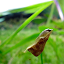} &
    \includegraphics[width=45px]{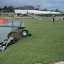} &
    \includegraphics[width=45px]{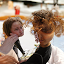} &
    \includegraphics[width=45px]{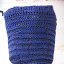} &
    \includegraphics[width=45px]{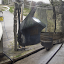} &
    \includegraphics[width=45px]{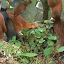} &
    \includegraphics[width=45px]{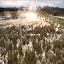}
    \\
    \includegraphics[width=45px]{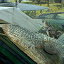} &
    \includegraphics[width=45px]{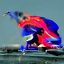} &
    \includegraphics[width=45px]{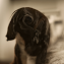} &
    \includegraphics[width=45px]{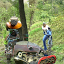} &
    \includegraphics[width=45px]{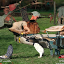} &
    \includegraphics[width=45px]{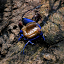} &
    \includegraphics[width=45px]{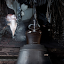} &
    \includegraphics[width=45px]{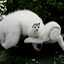}
    \\
    \includegraphics[width=45px]{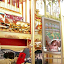} &
    \includegraphics[width=45px]{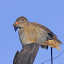} &
    \includegraphics[width=45px]{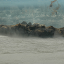} &
    \includegraphics[width=45px]{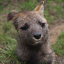} &
    \includegraphics[width=45px]{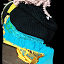} &
    \includegraphics[width=45px]{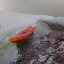} &
    \includegraphics[width=45px]{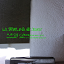} &
    \includegraphics[width=45px]{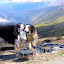}
    \\
    \end{tabular}
    \caption{Full batch of generated samples from the model trained on ImageNet using \textbf{1024 latents} during inference, the same as the number of latents used during model training.}
    \label{fig:imagenet-samples-z1024}
\end{figure*}

\begin{figure*}[t]
    \centering
    \begin{tabular}{cccccccc}
    \includegraphics[width=45px]{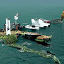} &
    \includegraphics[width=45px]{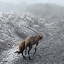} &
    \includegraphics[width=45px]{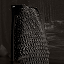} &
    \includegraphics[width=45px]{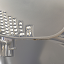} &
    \includegraphics[width=45px]{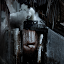} &
    \includegraphics[width=45px]{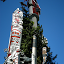} &
    \includegraphics[width=45px]{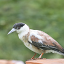} &
    \includegraphics[width=45px]{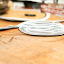}
    \\
    \includegraphics[width=45px]{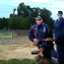} &
    \includegraphics[width=45px]{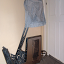} &
    \includegraphics[width=45px]{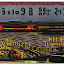} &
    \includegraphics[width=45px]{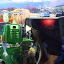} &
    \includegraphics[width=45px]{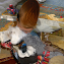} &
    \includegraphics[width=45px]{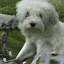} &
    \includegraphics[width=45px]{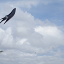} &
    \includegraphics[width=45px]{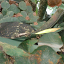}
    \\
    \end{tabular}
    \caption{Full batch of generated samples from the model trained on ImageNet using \textbf{1536 latents} during inference. The same random seed was used as when generating the images with 1024 latents, which explains how the image of the white dog appears in both batches.}
    \label{fig:imagenet-samples-z1536}
\end{figure*}

\begin{figure*}[t]
    \centering
    \begin{tabular}{cccccccc}
    \includegraphics[width=45px]{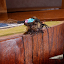} &
    \includegraphics[width=45px]{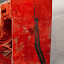} &
    \includegraphics[width=45px]{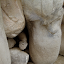} &
    \includegraphics[width=45px]{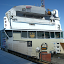} &
    \includegraphics[width=45px]{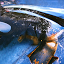} &
    \includegraphics[width=45px]{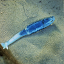} &
    \includegraphics[width=45px]{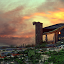} &
    \includegraphics[width=45px]{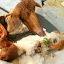}
    \\
    \includegraphics[width=45px]{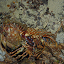} &
    \includegraphics[width=45px]{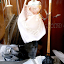} &
    \includegraphics[width=45px]{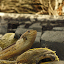} &
    \includegraphics[width=45px]{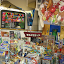} &
    \includegraphics[width=45px]{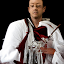} &
    \includegraphics[width=45px]{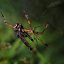} &
    \includegraphics[width=45px]{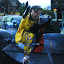} &
    \includegraphics[width=45px]{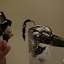}
    \\
    \end{tabular}
    \caption{Full batch of generated samples from the model trained on ImageNet using \textbf{2048 latents} during inference.}
    \label{fig:imagenet-samples-z2048}
\end{figure*}

\begin{figure*}[t]
    \centering
    \begin{tabular}{cccc}
    \includegraphics[width=96px]{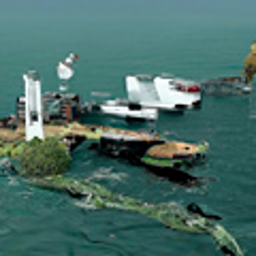} &
    \includegraphics[width=96px]{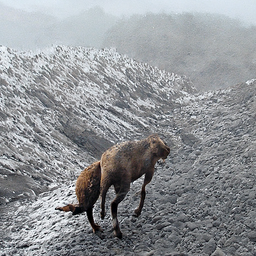} &
    \includegraphics[width=96px]{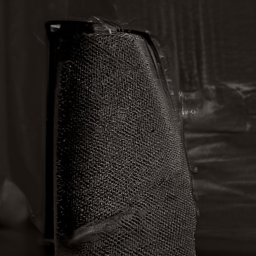} &
    \includegraphics[width=96px]{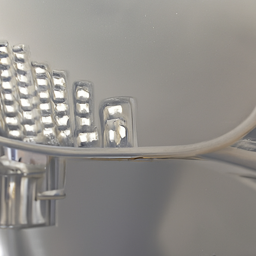}
    \\
    \includegraphics[width=96px]{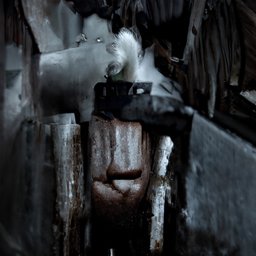} &
    \includegraphics[width=96px]{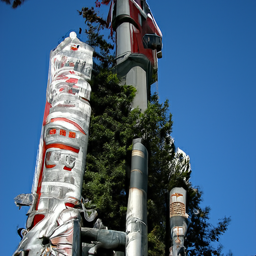} &
    \includegraphics[width=96px]{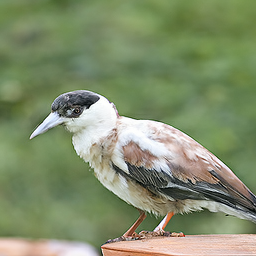} &
    \includegraphics[width=96px]{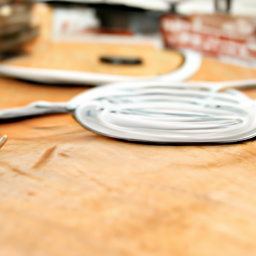}
    \\
    \includegraphics[width=96px]{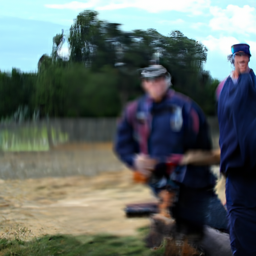} &
    \includegraphics[width=96px]{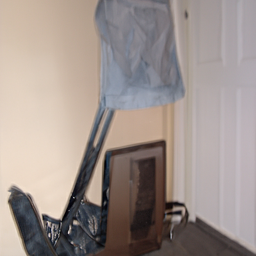} &
    \includegraphics[width=96px]{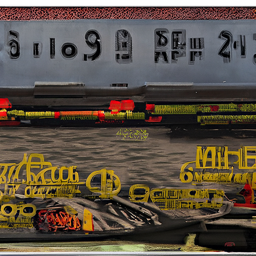} &
    \includegraphics[width=96px]{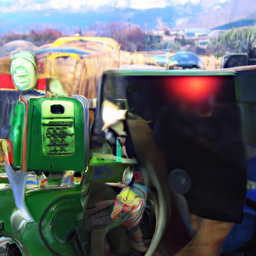}
    \\
    \includegraphics[width=96px]{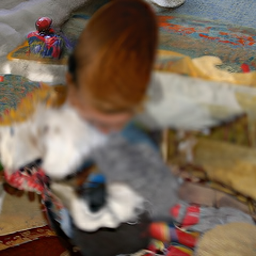} &
    \includegraphics[width=96px]{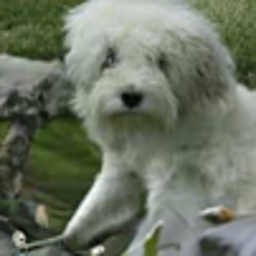} &
    \includegraphics[width=96px]{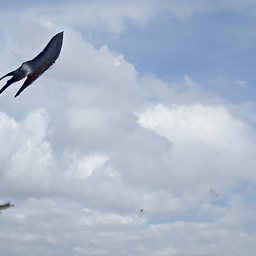} &
    \includegraphics[width=96px]{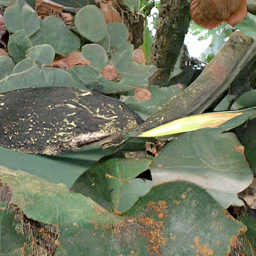}
    \\
    \end{tabular}
    \caption{Images from the batch generated with 1536 latents (\cref{fig:imagenet-samples-z1536}) upsampled to 256~$\times$~256 pixels using SR3 \cite{saharia2021image}.}
    \label{fig:imagenet-samples-z1536-super-res}
\end{figure*}

Full batches of the generated images used to populate \cref{fig:imagenet-samples} can be seen in \crefrange{fig:imagenet-samples-z16}{fig:imagenet-samples-z2048}. All images were generated with a temperature of $1.0$. We also show upsampled versions of the images that were generated with 1536 latents by using SR3 \cite{saharia2021image} to achieve a resolution of 256~$\times$~256 in \cref{fig:imagenet-samples-z1536-super-res}.

\section{Books}
\label{sec:books_appendix}
\begin{figure}[H]
    \centering
    \includegraphics[width=0.9\columnwidth]{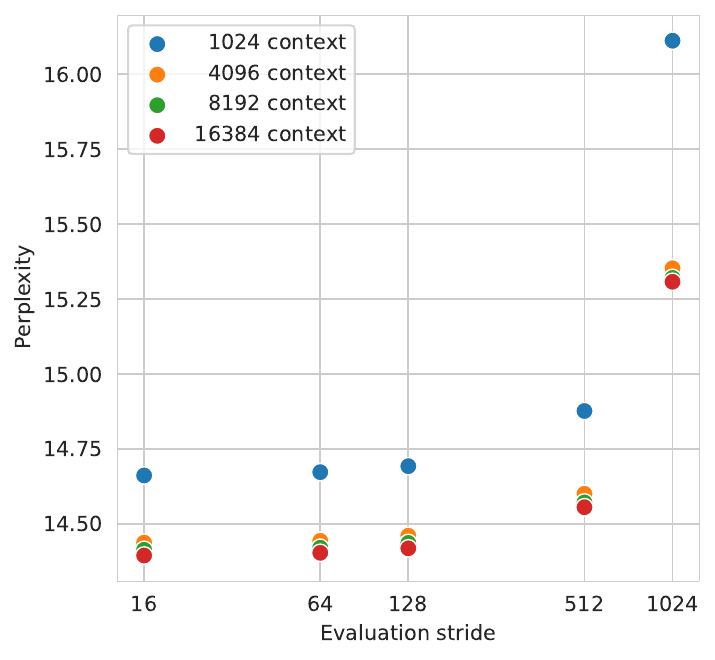}
    \caption{Perplexity results on the Books test set, from 4 different 36-layer Perceiver AR models with 1024-, 4096-, 8192- and 16384-contexts, respectively. The evaluation is done for 5 stride values.}
    \label{fig:books_eval_stride_vs_ppl}
\end{figure}

As discussed in \cref{sec:books}, we train all models on the Books dataset with 1024 latents, 36 layers and $\{1024, 4096, 8192, 16384\}$ input context tokens. In addition to the results shown in the main paper at stride 512, we also evaluate performance on our test set of 100 books as a function of stride (\cref{sec:eval_details}), looking at 5 values: $\{16, 64, 128, 512, 1024\}$ (\cref{fig:books_eval_stride_vs_ppl}). 

We draw the reader's attention to two effects here. \textbf{First}, while the perplexity at a given context length is relatively stable for strides $\le 128$, perplexity consistently increases with stride. When using strided evaluation, the first tokens in a model's context window see a relatively small number of preceding tokens. With an evaluation stride of 1024, the first token in the context window of a 1024-context model sees only one preceding token. This property is likely responsible for the increasing gap between 1024-context models and larger-context models as the stride increases: the perplexity gain moving from the perplexity gain of the 16384- over the 1024- model is 0.8 at stride 1024, but only 0.26 at stride 16. 

\textbf{Second}, regardless of the evaluation stride, each twofold increase in context improves perplexity, but with diminishing returns: the gap between 1024 and 4096 context is consistently larger than that between 4096 and 8192 or 8192 and 16384 contexts. Perceiver AR models with the same depth and different context sizes use the same number of parameters (with the possible exception of differences in the parameters of the position encoding). We believe this effect points to the need for larger capacity models to exploit the increased information in longer contexts. Nonetheless, the overall trend suggests that larger context leads to improved results, even when using essentially the same model capacity.

\section{A More Detailed Look at Perceiver AR's Internals}
\label{sec:perceiver_internals}

Like Perceiver and Perceiver IO, Perceiver AR is built on Transformer-style attention blocks. Perceivers use two types of attention: cross- and self-attention. These two types of attention share an interface---both take in two arrays and return a third---but differ in what they pass to that interface.

\textit{Zooming in:} QKV attention takes in a key-value input $X_{KV} \in \mathbb{R}^{M \times C}$ and a query input $X_{Q} \in \mathbb{R}^{N \times D}$ (where $C$ and $D$ indicate number of channels). The output of QKV attention is an array with the same index (first) dimension as the query input and a channel (second) dimension determined by an output projection:

\begin{align}
    Q = f_{Q}(X_Q) \text{;} \; K = f_K(X_{KV})  \text{;} \; V = f_V(X_{KV}) \\
    X_{QK}^{\text{pre}} = QK^T \\
    X_{QK} = \text{softmax}(X_{QK}^{\text{pre}} / \sqrt{F}) \\
    \text{Attn}(X_Q, X_{KV}) = X_{QKV} = f_{O}({X_{QK}V}),
\end{align}

where $X_{QK}^{\text{pre}}$ and $X_{QK}$ are the array of pre- and post-softmax attention maps $\in \mathbb{R}^{N \times M}$, and $X_{QKV}$ is an array $\in \mathbb{R}^{N \times D}$. The functions $f_{\{Q, K, V\}}$ are linear layers mapping each input to a shared feature dimension $F$ and $f_O$ is a linear layer projecting the output to a target channel dimension $D$, which is often the same size as $X_{Q}$'s. All linear layers are applied convolutionally over the index dimension. We have omitted batch and head dimensions (in the case of multi-headed attention) for readability.

In Perceiver AR attention blocks, QKV attention is followed by a two-layer MLP with a squared ReLU nonlinearity following the first layer. The full module has the following structure:

\begin{align}
    \label{eq:attend_eq_2}
    X_{QKV} &= \text{Attn}(\text{layerNorm}(X_Q),     
    \text{layerNorm}(X_{KV})) \\
    \label{eq:attend_eq_1}
    X_{QKV} &= X_{QKV} + X_Q \\
    \label{eq:attend_eq_3}
    X_{QKV} &= X_{QKV} + \text{MLP}(\text{layerNorm}(X_{QKV})),
\end{align}

slightly abusing notation for simplicity and to emphasize the residual structure. ``Attn'' refers to QKV as described above.

\textit{Zooming out:} When discussed in the main text, the operations $\text{CrossAttend}: X_{KV} \times X_{Q} \rightarrow X_{Q}$ and $\text{SelfAttend}: X_{KV} \times X_{Q} \rightarrow X_{Q}$ refer to the full system of equations given in \cref{eq:attend_eq_1,eq:attend_eq_2,eq:attend_eq_3}. 

These two operations differ only in that $X_Q \neq X_{KV}$ for cross-attention (with $N < M$ for the ``encoder'' cross-attention considered here) and $X_Q = X_{KV}$ for self-attention. Cross-attention is used to reduce the shape of an input array and self-attention to keep it the same shape. In Perceiver and Perceiver IO, the initial cross-attention's query input $X_{Q}$ is typically learned (its elements are ``learned latents''), while in Perceiver AR, it is typically constructed\footnote{With the exception of experiments on Wikitext-103, where learned latents are used.} by taking the last $N$ elements of the input array: $X_{Q} = X_{KV}[-N{:}, :]$, using NumPy-style indexing notation \cite{harris2020array}. In either case, using fewer latents than inputs is essential to controlling compute and memory costs while keeping long-context inputs.

Perciever AR also differs from Perceiver and Perceiver IO in the use of causally masked cross- and self-attention. In self-attention masks, all elements of $X_{QK}^{\text{pre}}$ at indices $(m', m)$ (queries $\times$ keys), where $m', m \in [0, M)$ and $m > m'$ are masked. In the cross-attention mask, to compensate for the fact that latents are placed at the trailing index locations, all elements of $X_{QK}^{\text{pre}}$ at indices $(n, m)$ (queries $\times$ keys), where $n \in [0, N), m \in [0, M)$ and $m > n + M - N - 1$ are masked. This prevents ``earlier'' queries from attending to ``later'' keys. Causal masking is implemented in attention by multiplying all masked connections in the pre-softmax attention map $X_{QK}^{\text{pre}}$ by $-\infty$. 

If we denote causally masked self- and cross-attention by $\text{CrossAttend}_{\text{cm}}$ and $\text{SelfAttend}_{\text{cm}}$, respectively, Perceiver AR (without learned latents\footnote{For latents, replace the second (query) input to the RHS of \cref{eq:cross_attend_eq} with $Z_0$ (or $Z_{-1}$ if you like), which is learned.}) is given by the following:

\begin{align}
\label{eq:cross_attend_eq}
    Z_0 \leftarrow \text{CrossAttend}_{\text{cm}}(X, X[-N{:},:])
\end{align}

\begin{align}
\label{eq:self_attend_eq}
    Z_{l+1} \leftarrow \text{SelfAttend}_{\text{cm}}(Z_l, Z_l),
\end{align}

where \cref{eq:self_attend_eq} is applied once per self-attend layer. The final output is obtained by layer-norming and projecting ${Z_{L}}$ to the vocabulary size, followed by a softmax to produce output logits.

\section{Further Related Work}
\label{sec:further_related_work}

In this section we describe additional background with the goal of elucidating Perceiver AR's problem setting and method.

\subsection{Efficient attention}
\label{sec:efficient_attention}
\noindent
Many recent Transformer variants seek to avoid the $O(N^2)$ memory requirements of self-attention. This is often done by introducing sparsity when computing the attention matrix  -- as in Sparse Transformer \cite{child2019generating}, Big Bird \cite{zaheer2020bigbird}, and Combiner \cite{ren2021combiner} -- or by approximating this computation at lower cost -- e.g. as in Linear Transformer \cite{katharopoulos2020transformers}, Linformer \cite{wang2020linformer}, Reformer \cite{kitaev2020reformer}, Random-Feature Attention~\cite{peng2021random}, and Performer \cite{choromanski2021rethinking}.

The downside of methods that use sparsity is that this sparsity must be hand-tuned or created with heuristics that are often domain specific and can be hard to tune. In contrast, our work does not force a hand-crafted sparsity pattern on attention layers, but rather allows the network to learn which long-context inputs to attend to and propagate through the network. The initial cross-attend operation, which reduces the number of positions in the sequence, can be viewed as a form of learned sparsity.

Because Perceiver AR does not depend on hand-tuned sparsity patterns or e.g. structured dilation patterns like those used in WaveNet \cite{oord2016wavenet}, it can model arbitrarily complex dependency patterns between any of its inputs immediately after the cross-attend. Models with fixed sparsity patterns, on the other hand, typically require several layers (which depends logarithmically on the distance between points in the input array in the case of WaveNet) or precise and fragile conjunctions of input receptive fields (in the case of hand-tuned sparsity patterns) to allow a given set of points to interact. The net effect of this situation is that the effective processing used to process a given set of features is much smaller in these architectures than densely connected architectures like Perceiver AR.

\subsection{Other efficient architectures}
\label{sec:other_efficient_archs}
\noindent
\textbf{Memory-based models.}
Longer effective context size can also be achieved by reducing the compute requirement on tokens that are far in the past using stop gradients \cite{dai2019transformerxl}, recurrence \cite{mehri2017samplernn}, memory \cite{weston2014memory,rae2019compressive}, or other forms of compression \cite{wu2022memorizing} to process long-term structure. These strategies typically impose bottlenecks with local structure, which limits the flexibility with which context can be exploited by a given target. Although Perceiver AR performs processing using latents that are fewer in number than the inputs, each latent is given direct access to all inputs, rather than communicating with the past through a narrow or precomputed mechanism.

\noindent
\textbf{Alternatives to attention.}
Efficiency can also be obtained using domain-tuned architectures such as lightweight or dynamic convolutions~\cite{wu2019pay}.
The recently introduced S4 \citep{gu2021s4} is a very efficient model that avoids attention altogether while producing interesting results, but it is not yet competitive on standard datasets like WikiText-103.

Many of the insights presented in this prior work are complementary to Perceiver AR's architectural design, and we expect they can be hybridized to produce even more efficient models suitable for long-scale, general purpose model design in future work.

\subsection{Input Tokenization}
\label{sec:input_tokenization}
Another approach to reducing the memory and compute requirements of self-attention is to directly reduce the length of the input data by using tokenization to group multiple inputs into single tokens. Such approaches have led to excellent results on a variety of domains of interest and are often an implicit feature of the data arrays in standard datasets.

In NLP, subword chunks are often grouped together~\cite{kudo2018sentencepiece} based on their frequency in a training text corpus. In vision, previous work has explored using K-means to group RGB values into a single token~\cite{chen2020generative} or to group individual pixels into patches~\cite{dosovitskiy2020image}, sometimes followed by quantization~\cite{ramesh2021zero}. Others have also leveraged DCT coefficients~\cite{nash2021generating} as used in JPEG to convert fixed-size images to variable-length sequences. Arguably the most widely applied tokenization method is neural vector quantization, where an encoder network is trained to map images to a spatially downsampled collection of discrete codes~\cite{oord17discrete, ramesh2021zero}.
Similar techniques have been developed for audio, where vector-quantized encodings of raw waveforms have effectively been used as vocabularies for end-to-end speech synthesis~\cite{lakhotia2021generative, polyak2021speech} and music generation~\cite{dhariwal2020jukebox}.
The SoundStream codec we use in this paper builds upon these techniques by using residual vector quantization and adversarial reconstruction losses to achieve high audio fidelity with fewer discrete tokens~\cite{zeghidour2021soundstream}.

Tokenization is a broadly useful strategy, but it has its downsides. For many domains, effective tokenization schemes rely on lossy compression. Data is discarded in the tokenization process, and inputs can not be recovered exactly. Care is required to ensure that the data can be reconstructed at a fidelity adequate for the required application. Neural compression schemes such as VQ-VAE require users to train and maintain additional encoder and decoder networks, which can hinder ready application to new datasets and domains. And, perhaps most tellingly, effective tokenization is typically designed and used in a domain-specific fashion, which limits the ease of adaption and scaling to new domains. By effectively modeling long sequences, PerceiverAR can in some cases eliminate the need for tokenization and in others can reduce the need for heavy lossy compression. But in the near term, we anticipate that tokenization --- of one form or another --- will remain a necessary tool for incorporating more context. 

\section{Additional Details of the Methods}
\label{sec:additional_details}

\begin{figure*}
    \centering
    \includegraphics[width=\textwidth]{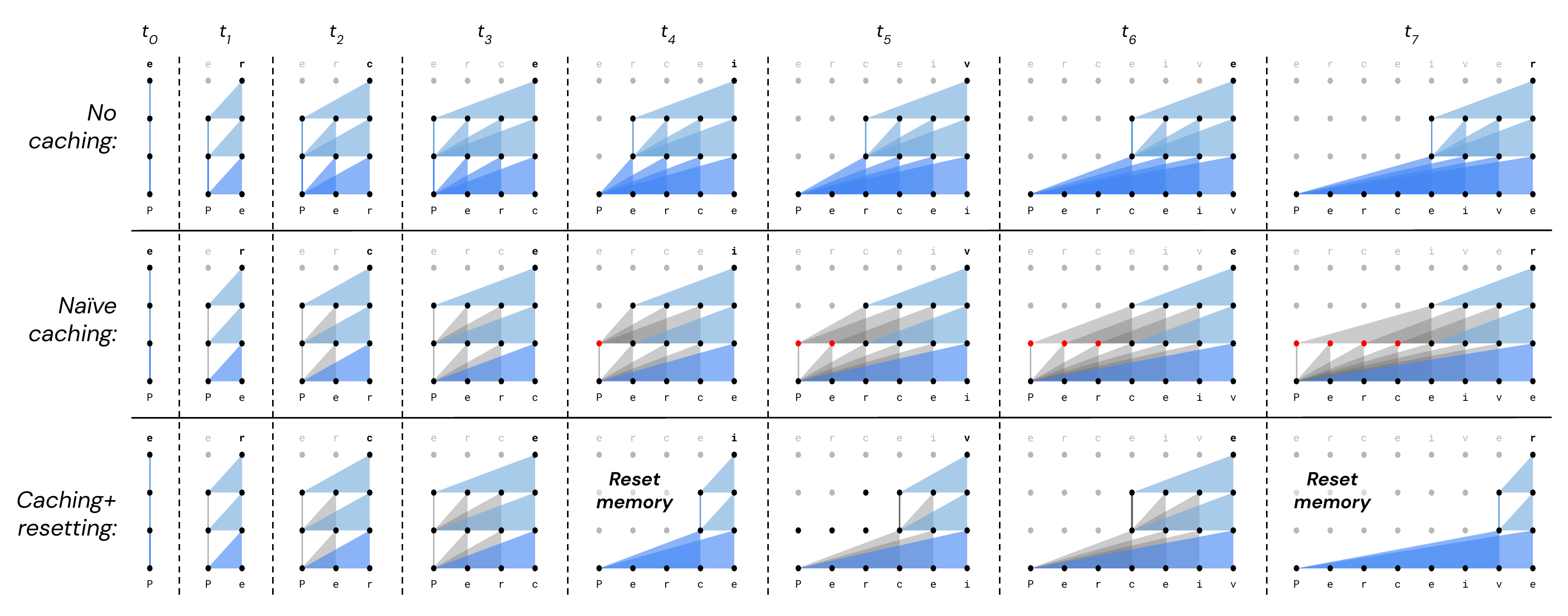}
    \vspace{-10pt}
    \caption{\textbf{Figure best viewed on a screen.} Caching at generation time allows previously computed states to be reused but introduces long-term dependencies not seen at training time when the input size and latent size differ. We illustrate this effect here for a Perceiver AR with $N=4$ latents and an input context $M=8$. Here, blue triangles indicate which attention operations are performed in the current step, while gray triangles indicate reused (cached) computations. The top row (no caching) matches what happens at training time. If caching is applied naively (middle row), the amount of computation per step can be greatly reduced. However, as the model is run out for more steps than there are latents, caching introduces dependencies on latents that are no longer active but have already been used to compute latents that are active. These latents are shown in red. In other words, caching previous latents allows longer-distance latents (which are no longer active) to influence the current generation, introducing long-term dependencies that were not encountered at train time. We find in practice that these long-term dependencies lead to degraded performance when caching is run out for too many steps. To avoid this problem, we cache by periodically resetting memory, allowing some computation to be reused but avoiding the introduction of long-term dependencies not encountered at training time. This strategy is described in detail in \cref{sec:activation_caching_inference}.
}
    \label{fig:caching_long_term_dependencies}
\end{figure*}

\subsection{Memory Usage}
\label{sec:memory_usage}
The single largest source of memory usage in a Perceiver AR model is typically the attention map in the initial cross-attend layer, which results in a matrix of size $[\mathrm{heads}, \mathrm{input\_length}, \mathrm{self\_attention\_length}]$. For experiments in this paper where this matrix caused out of memory errors (\cref{sec:copy_task_input_length}), we found that processing attention heads in subgroups rather than all at once was sufficient to reduce our memory usage. Using the chunked approach in \cite{kitaev2020reformer, rabe2021self, jumper2021highly} for the cross-attend layer allows scaling input length beyond even those limits without requiring any architecture changes or adding compute requirements for layers other than the cross-attend. These memory-saving tricks do result in reductions to training throughput (in steps per second), so we avoid them where possible.

\subsection{Cross-attend Dropout}
\label{sec:cross_attend_dropout}

Dropout \cite{JMLR:v15:srivastava14a} is used by default in many Transformer implementations and is an essential tool for mitigating overfitting on small datasets. Dropout is typically imposed on linear layers after the attention softmax or within the Transformer MLP block. Perceiver AR supports this kind of dropout, but the cross-attend layer also enables interesting possibilities for dropout before the attention softmax.

We find that for some tasks, masking out positions in the initial cross-attend is an effective way of preventing overfitting. This can also be used to save memory by setting a budget for a certain number of inputs and then selecting that number of inputs randomly from the maximum input context. Because no position-specific parameters are learned for the cross-attend layer, a smaller number of inputs can be used at train time than during evaluation or inference.

Imposing cross-attend dropout can be interpreted as enforcing high sparsity at training time, but allowing less extreme sparsity at evaluation time. Because the attention layer itself is scale invariant, the uniform scaling normally imposed by dropout at train time is unnecessary.

\subsection{Activation Caching for Inference}
\label{sec:activation_caching_inference}

Naively sampling from a Transformer for inference can be very slow because activations for all positions must be calculated at every step. Caching the key/value activations for previously inferred positions is a common technique for improving generation speed \cite{tensor2tensor,shazeer2019fast}. Perceiver AR can use a similar approach, but the exact technique cannot be applied directly because the number of output positions is smaller than the number of input positions, so preserving all previous activations would result in an effective self-attention stack as wide as the number of inputs.

Transformer-XL \cite{dai2019transformerxl} solves this problem by keeping a buffer of activations for only the last N positions. This works because the model is presented with activations for that number of previous positions during training, which is not the case for Perceiver AR. We also cannot simply restrict the buffer size to be the same as the number of targets because even when activations for a given position are expired, they have already influenced other positions within the buffer (\cref{fig:caching_long_term_dependencies}).

Instead, we apply a simple trick to ensure that no cached activations are influenced by positions beyond what was seen at training time. We use a fixed activation buffer the same width as the self-attention stack at train time. When the buffer is full, we do a full forward pass without any cached activations for the next position, but using half the number of latents. The activations from that pass are saved in the buffer, leaving it half full. Inference then proceeds until the buffer is full again. The activations from the cross-attend layer do not require this trick unless inference will extend beyond the length of the inputs used at train time.

We find that these occasional full forward passes add minimal overhead, and the speed gains from using cached activations are still significant. We performed a test inferring a single image with a sequence length of 12,289 using a model with 1024 latents (see \cref{sec:imagenet-64x64} for details) on a single TPUv3 core to compare speeds. Inference without any caching took 7.93 minutes. With caching the same task took 3.68 minutes, less than half the time.

\subsection{Varying Compute at Test Time}
\label{sec:varying_compute_at_test_time_appendix}

\begin{figure}[h]
    \centering
    \includegraphics[width=.9\columnwidth]{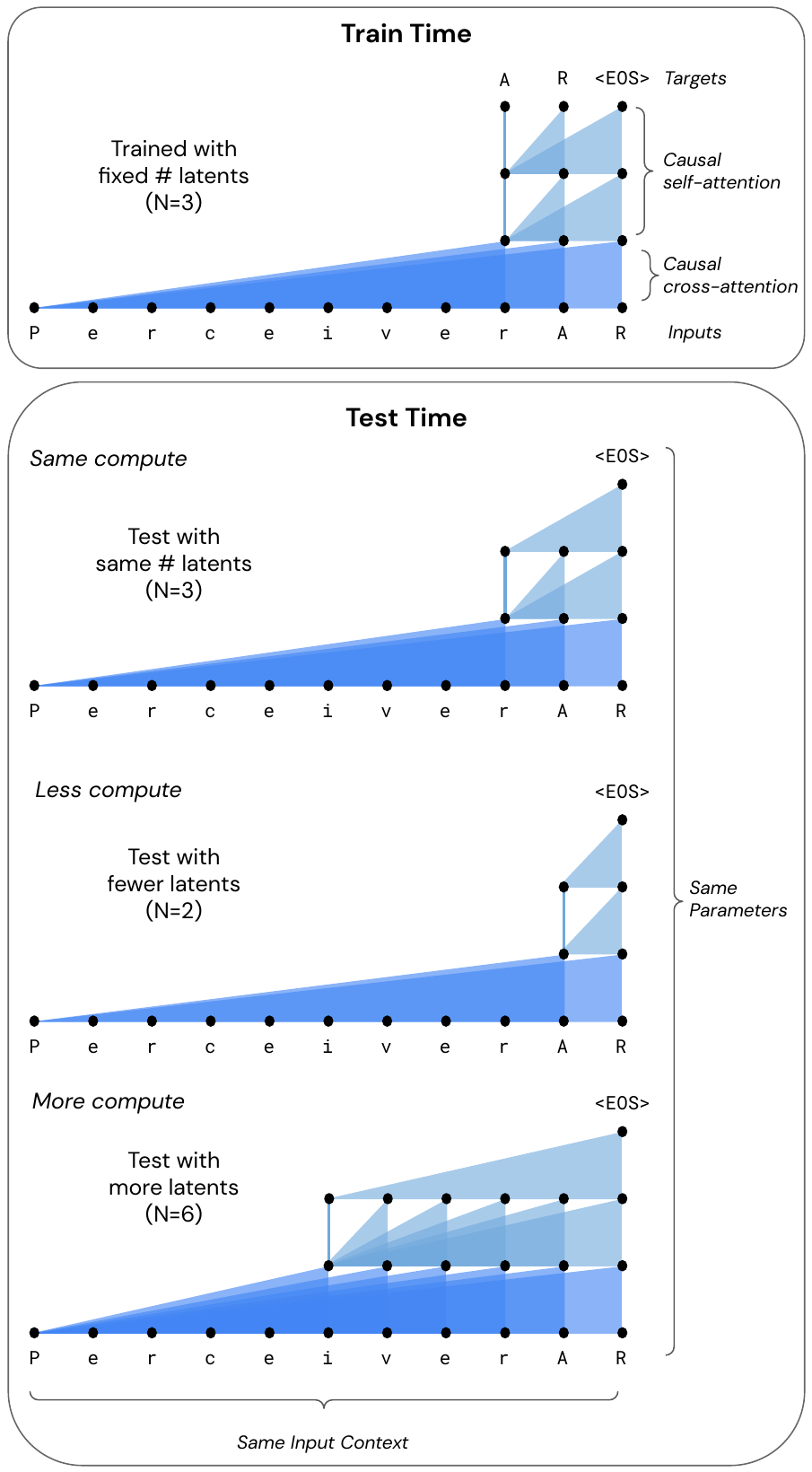}
    \caption{Because Perceiver AR decouples input length from the width of the self-attention stack, the number of latents can be different at train time and test time. This does not require additional training because no per-position parameters are learned in the self-attention stack. \Cref{sec:compute_at_test_time} discusses how this possibility can be used to scale up or down compute requirements and output quality.}
    \label{fig:test_time_configs}
    \vspace{-20pt}
\end{figure}

\Cref{fig:test_time_configs} illustrates how the number of latents can be changed at test time without changing either the trained parameters of the model or the model's input context length. In \cref{sec:compute_at_test_time} we discuss how this possibility can be used to scale up or down compute requirements and output quality.

\section{Training Details}
\label{sec:training_details}

\subsection{Common}

Unless otherwise stated, models were trained with the following configuration.

We use the Adam optimizer \cite{kingma2015adam} as implemented in the Optax framework \cite{optax2020github} with $b1=0.1$, $b2=0.999$, $eps=1\mathrm{e}{-8}$, a base learning rate of $3\mathrm{e}{-4}$, and a 10k step linear warmup. To reduce memory usage, we used ZeRO Stage 1 optimizer state partitioning \cite{Rajbhandari2020ZeRO}.

For training stability, we used a global max norm of $1.0$. We also added an additional loss term 
  $z\_loss*log(z)^2$ with $z\_loss=1\mathrm{e}{-4}$, as used in training the T5 family of models \cite{raffell2020exploring}.
 
Both the input embedding and latent vectors were size 1024, and 16 attention heads were used for the initial cross-attend and within the self-attention stack. Within the MLP layers of the cross-attend and self-attention stack, the input dimensionality is projected to 4x its size and Squared ReLU activations \cite{so2021primer} were used. We used a cross-attend dropout probability of $0.1$. 

Training and evaluation were done on either TPUv2 or TPUv3 clusters.

\subsection{Rotary Position Encodings}
We encode the position of tokens using rotary position encoding \cite{su2021roformer}. With this method, rotation matrices (built from sine and cosine functions, just as with sinusoidal/Fourier position encodings) rotate pairs of dimensions in each of the key and query heads to reflect the absolute position of the key or query in the sequence. When used to compare a given key and query pair, these rotations produce attention weights that reflect only the relative distance between tokens. The result is a memory-efficient relative position mechanism.

\begin{figure}[h]
    \centering
    \includegraphics[width=.85\columnwidth]{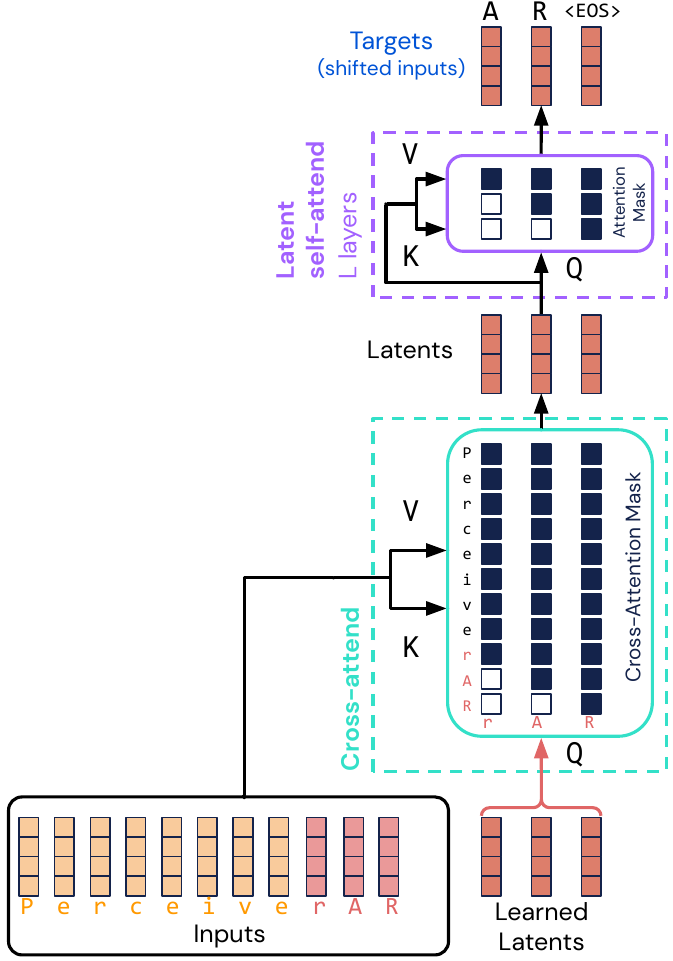}
    \caption{Variant of Perceiver AR where we feed learned input latents instead of the input embeddings directly. This is used for our Wikitext-103 experiments as we noticed it helped reduce overfitting.}
    \label{fig:perceiver_ar_architecture_with_learned_latents}
    \vspace{-10pt}
\end{figure}

Intuitively, this mechanism represents position using a strategy somewhat like that used by analogue clock hands. Like a clock hand,  each pair of dimensions (sine/cosine) in the rotary position encoding is responsible for indicating position at some frequency (second, minutes, hours, etc.). When we have multiple clock hands (multiple dimensions within each token vector), each of which moves at a different frequency, we can precisely tell both the current time (q/k position) and we can resolve it at different resolutions, depending on what's needed for a given task.

We found in early experiments that rotating a fraction of channel dimensions led to better results, which we discovered has also been noticed (the first 50\%) \cite{kingoflolz2021github}. Fractional rotation of this kind reduces the fidelity with which we encode position information (because it results in fewer frequency bands). The result of fractional rotation is that only some channel dimensions are modulated by the relative position between a query and key. This may encourage the network to exploit both position-dependent and position-agnostic relationships between queries and keys when computing attention weights. Prior work characterizing the effect of position on attention weights suggests that many common position encoding strategies bias the network towards attending to recent tokens and fractional rotation may mitigate this effect. Fractional rotation also but makes the rotation computation significantly cheaper, as it requires fewer matrix multiplies.

\subsection{Copy Task}
\label{sec:copy_task_supmat}

The copy task was trained with 1024 latents in a 6-layer self-attention stack. For position encoding, we used fixed sinusoidal embeddings to denote absolute position, as described in the original Transformer paper \cite{vaswani2017attention}. There were 4096 sequences per batch.

To reduce the instantaneous memory requirements created by attending to the input sequence of length 131,072, the 16 cross-attention heads were split into 4 groups of 4 and computed separately (see \cref{sec:memory_usage}).

The first 1K steps were a linear learning rate warmup and the remaining followed a cosine learning rate decay schedule.

\subsection{ImageNet 64~$\times$~64}

The model trained on ImageNet 64~$\times$~64 has 770.1M parameters. Training proceeded for a total of 750k steps. After an initial 10k step linear warmup, a constant learning rate of $3\mathrm{e}{-4}$ was used until the final 50k steps, which used a cosine decay to 0.

\subsection{Wikitext-103}
\label{sec:wikitext-103_appendix}

\begin{table*}
  \centering
  \begin{tabular}{c | c c | c c | c c}
    \toprule
    SoundStream bitrate & Context (32k) & Context (8k) & Test (32k) & Test (8k) & Validation (32k) & Validation (8k) \\
    \midrule
    12kbps & 27.2s & 6.8s & 2.31 & 2.25 & 2.28 & 2.27  \\
    18kbps & 18.4s & 4.6s & 2.52 & 2.53 & 2.51 & 2.52  \\
    22kbps & 14.8s & 3.7s & 2.55 & 2.60 & 2.55 & 2.60  \\
    \bottomrule
  \end{tabular}
  \caption{Perceiver AR negative log-likelihood results on SoundStream audio generation, from two models with context lengths of 8192 (8k) and 32768 (32k), respectively.} \label{tab:maestro_v3_ss_32k_vs_8k}
  \vspace{-10pt}
\end{table*}

We train 18-layer Perceiver AR models with 1,024 latents, adaptive inputs embeddings \citep{baevski2018adaptive}, and with increasing context length from 1,024 to 8,192 tokens.
One key difference to other experiments is that we use learned input latents rather than the input embeddings as illustrated in Figure~\ref{fig:perceiver_ar_architecture_with_learned_latents}.
We observe this variant to help reduce overfitting.
We also apply cross-attend dropout (\cref{sec:cross_attend_dropout}) with $p{=}0.15$ to the context tokens for which we predict the next token.

The Transformer-XL baseline used a memory length of 384 at train time and 1600 at eval time, for a full effective context length of 2624. We used a memory dropout rate of 0.25.

Perceiver AR models trained on Wikitext-103 have the following parameter counts: 356.5M (1024 context), 357.7M (2048 context), 359.8M (4096 context), 364.0 (8192 context). Note that the parameter count increases slightly with increasing context length because of the use of absolute position encodings on Wikitext-103. The baseline Transformer-XL has 285.2M parameters.

\subsection{PG-19}

Both Perceiver AR models trained on PG-19 use 974.6M parameters.

\subsection{Books}

All Books models reported in~\cref{tab:books_results} have 498.9M parameters.

Perceiver AR models reported in~\cref{tab:books_ar_36l_matched_by_txl_per_context} all have 498.9M parameters. Compute matched Transformer-XL models have the following parameter counts, by number of layers: 346.6M (23), 360.3M  (24), 373.9M (25), 414.8M (28).

Perceiver AR models reported in~\cref{tab:books_42l_txl_matched_by_ar_per_context} have the following parameter counts, by context length and number of layers: 826.4M (1024, 62L), 813.8M (4096, 61L), 801.2M (8192, 60L), 750.8M (16384, 56L). The 42-layer Transformer-XL with matched compute has 605.8M parameters.

We use the same memory settings for Transformer-XL baselines as on Wikitext-103: all Transformer-XL models have a full effective context length of $1600 + 1024 = 2624$.

\subsection{Music Generation Tasks}

All models use a value of $0.7$ for the cross-attend dropout described in \cref{sec:cross_attend_dropout}. All but the MAESTRO SoundStream-task models apply rotation to 25\% of the attention dimensions. Additionally, the models trained on the piano dataset and MAESTRO SoundStream tasks use a learning rate of $2\mathrm{e}{-4}$. The ones trained on the symbolic MAESTRO data use a learning rate of $1\mathrm{e}{-4}$, 1 initial cross-attend head and 4 heads within the self-attention stack. The model trained on the piano transcription dataset uses $0.1$ for cross-attend dropout and a $0.25$ dropout rate on the latents. Finally, this model pre-embeds the query inputs using a 3-layer MLP with GELU~\cite{hendrycks2016gelu} activations.

\begin{figure}[t]
    \centering
    \includegraphics[width=\columnwidth]{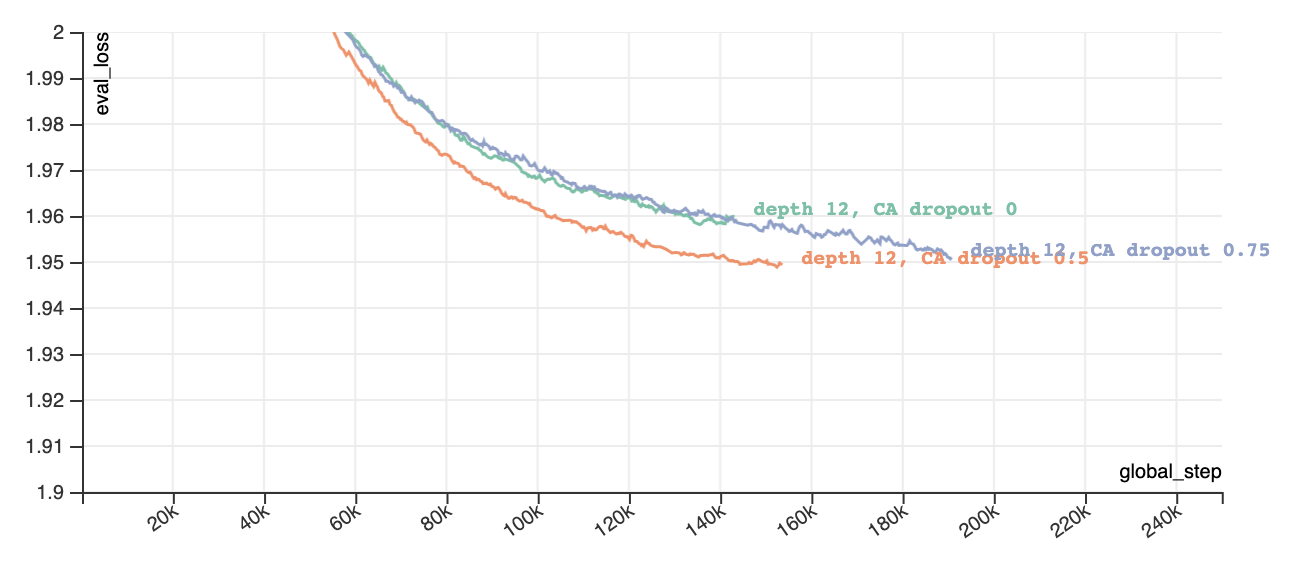}
    \includegraphics[width=\columnwidth]{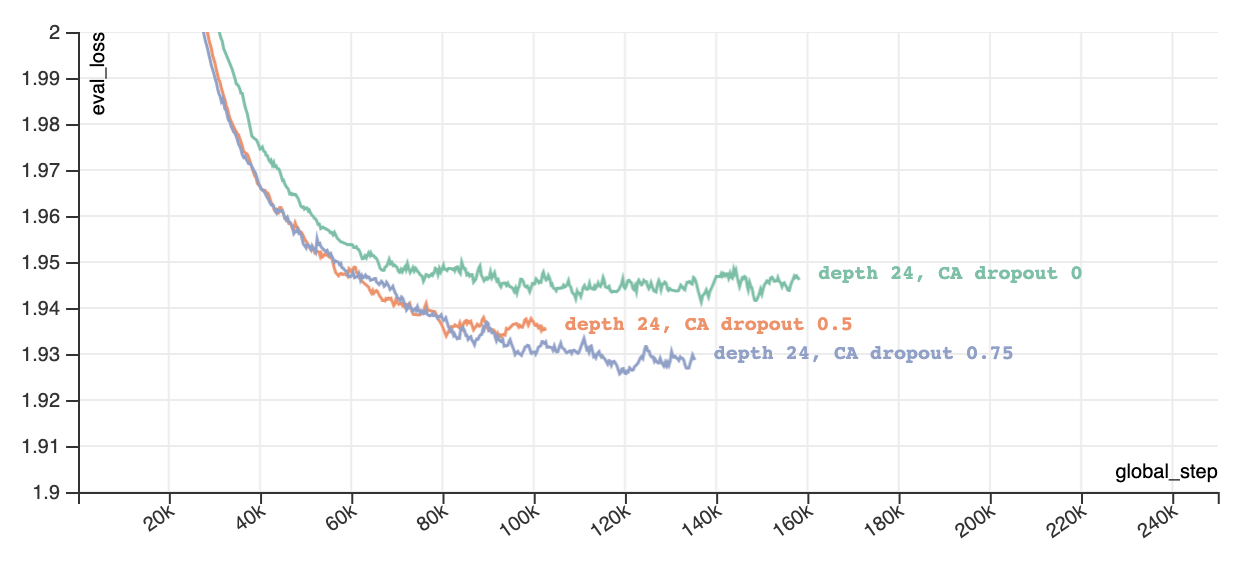}
    \includegraphics[width=\columnwidth]{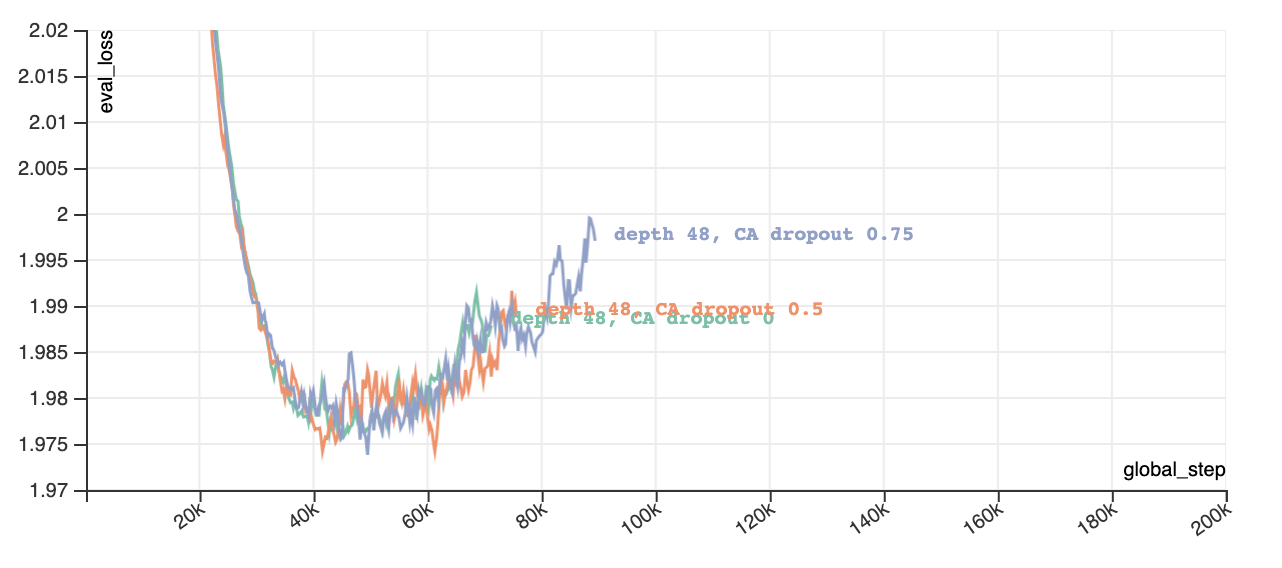}
    \vspace{-20pt}
    \caption{Validation loss on MAESTRO v3 as a function of the cross-attend dropout applied to the model. Each value $\{0, 0.5, 0.75\}$ is applied at 3 different model depths $\{12, 24, 48\}$.}
    \label{fig:ca_dropout_vs_depth_maestrov3}
\end{figure}

Audio from MAESTRO v3 was first processed using the SoundStream 12kbps codec. This bitrate of 12kbps corresponds to a vocabulary of 1024 tokens (10 bits), 24 tokens/frame, and 50 frames/second. This reconstructs audio while reasonable quality, compressing each second of audio (16k wave points) into 1.2k sequential tokens. Our trained model has an input context of length 8192 ($\sim$6.8 seconds) and uses 1024 latents in 12 self-attention layers. We also train and evaluate this configuration using higher-bitrate codecs---18kbps and 22kbps, with 36 and 44 tokens/frame respectively.

\section{Evaluation Details}
\label{sec:eval_details}
As discussed in Shortformer \cite{press2021shortformer}, there is a quality vs. speed tradeoff when evaluating long sequences with regard to stride. A stride of 1 (maximum overlap) is the slowest but highest quality, and a stride equal to the input length (no overlap) is the fastest but lowest quality. Because Perceiver AR decouples the number of targets from the number of inputs, our stride options range from 1 to the number of latents. We found that a stride of half the number of latents gave a good balance between speed and quality, and that is what we used for all the evaluations in this paper, unless otherwise mentioned. For a related set of issues for inference, see the discussion in \cref{sec:activation_caching_inference}.

\section{Dropout Ablations}

We study the effects of applying different forms of dropout to Perceiver AR, at several model depths when training on the MAESTRO v3 symbolic dataset.

Figure~\ref{fig:ca_dropout_vs_depth_maestrov3} illustrates the behaviour of the model when varying the amount of cross-attend dropout applied to it. The model has an input context of 4,096 and a post-attention dropout value of 0.5---both hyperparameters are kept constant across all runs. This type of dropout appears less useful when increasing model depth, up to having no effect at all during training for the 48-layer model.

\begin{figure}
    \centering
    \vspace{-10pt}
    \includegraphics[width=\columnwidth]{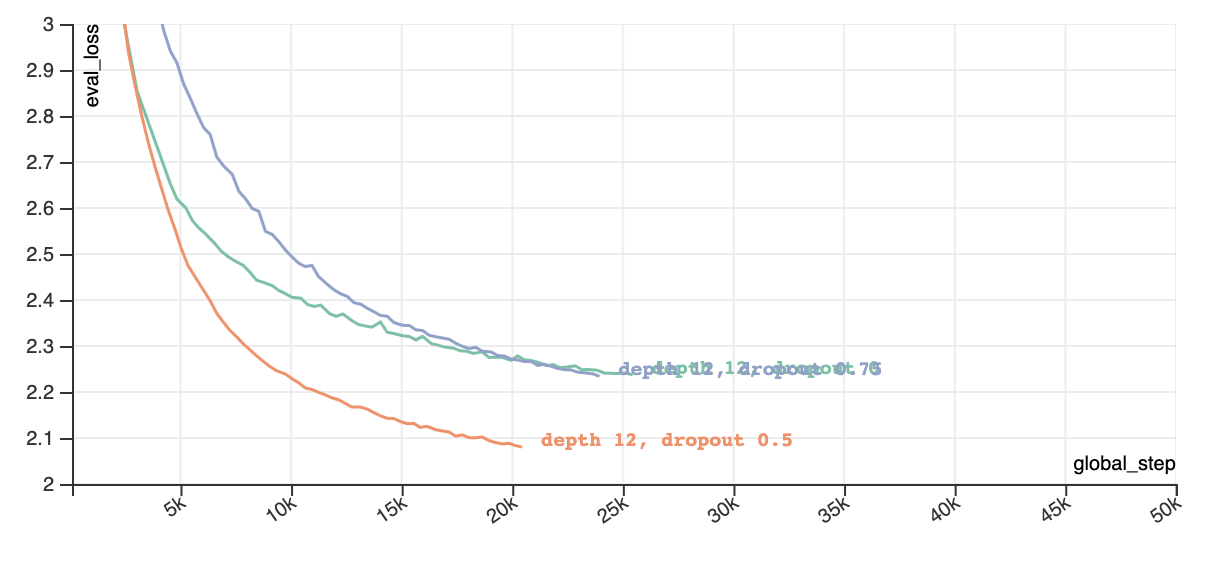}
    \includegraphics[width=\columnwidth]{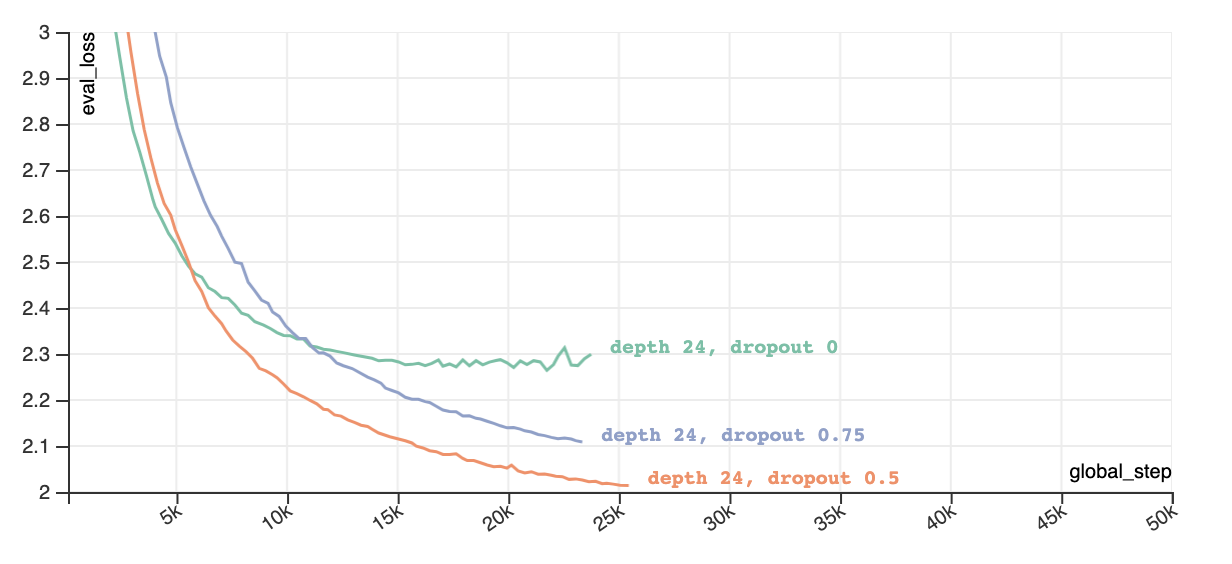}
    \includegraphics[width=\columnwidth]{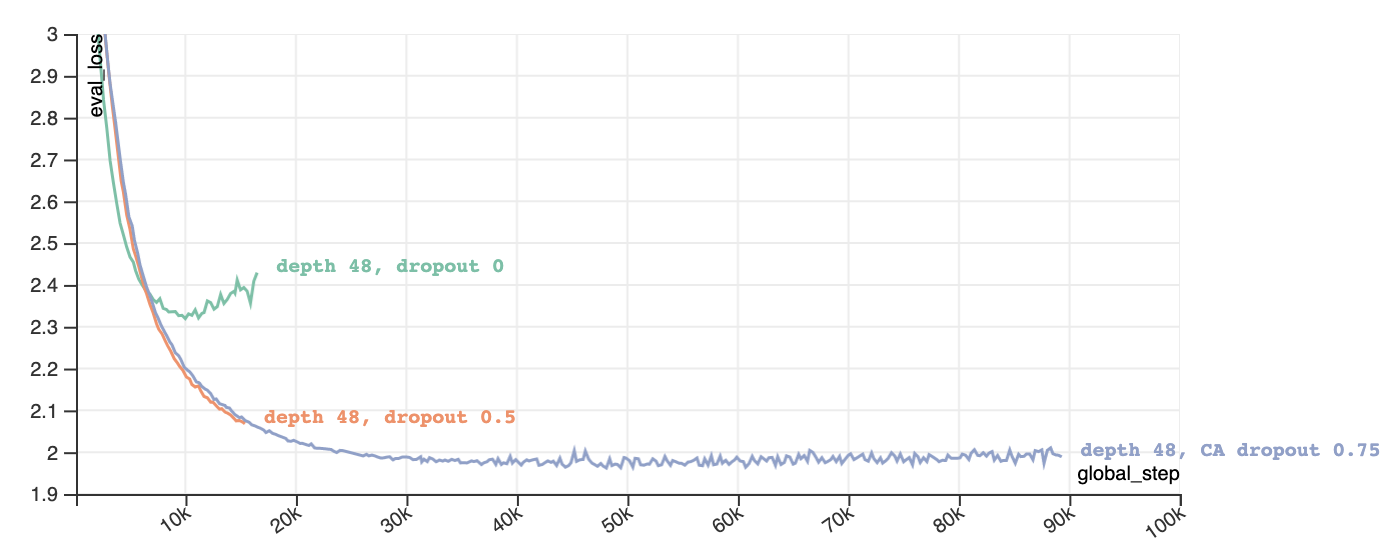}
    \vspace{-20pt}
    \caption{Validation loss on MAESTRO v3 as a function of the post-attention dropout applied to the model. Each value $\{0, 0.5, 0.75\}$ is applied at 3 different model depths $\{12, 24, 48\}$.}
    \label{fig:dropout_vs_depth_maestrov3}
\end{figure}

For the second experiment, we vary the amount of (post-attention) dropout in the same settings as previously described, while keeping a constant value of 0.7 for the cross-attend dropout. Figure~\ref{fig:dropout_vs_depth_maestrov3} shows that higher dropout rates become more useful at bigger depths---while at depth 12, applying 0.75 dropout actually hurts model training, this becomes beneficial at depth 48, where applying no dropout leads to quick overfitting.

\section{MAESTRO SoundStream}

In Table~\ref{tab:maestro_v3_ss_32k_vs_8k}, we compare results on all MAESTRO SoundStream tasks obtained from models that were trained on contexts of 8192 and 32768 tokens, respectively. While a smaller context length yields better results at the lowest bitrate, the 32k-context model obtains a significantly lower negative log-likelihood than the 8k- one on SoundStream 22kbps. Furthermore, the gap between corresponding NLLs widens as the bitrate increases. This result suggests that a shorter context is less and less useful as we attempt to produce higher-fidelity audio. The 32k context models reported here have the following parameter counts: 366.3M (12kbps), 391.5M (18kbps), 408.3M (22kbps). The 8k context models reported here have the following parameter counts: 215.2M (12kbps), 240.4M (18kbps), 257.1M (22kbps).

65536-context SoundStream models reported in~\cref{tab:maestro_v3_ss_results} have the following parameter counts: 668.6M (12kbps), 693.8M (18kbps), 710.6M (22kbps).

\end{document}

%% file: math_commands.tex
\usepackage{amsmath,amsfonts,bm}

\def\eqref#1{equation~\ref{#1}}

\def\1{\bm{1}}

\DeclareMathAlphabet{\mathsfit}{\encodingdefault}{\sfdefault}{m}{sl}
\SetMathAlphabet{\mathsfit}{bold}{\encodingdefault}{\sfdefault}{bx}{n}

%% file: main.bbl
\begin{thebibliography}{74}
\providecommand{\natexlab}[1]{#1}
\providecommand{\url}[1]{\texttt{#1}}
\expandafter\ifx\csname urlstyle\endcsname\relax
  \providecommand{\doi}[1]{doi: #1}\else
  \providecommand{\doi}{doi: \begingroup \urlstyle{rm}\Url}\fi

\bibitem[Baevski \& Auli(2018)Baevski and Auli]{baevski2018adaptive}
Baevski, A. and Auli, M.
\newblock Adaptive input representations for neural language modeling.
\newblock In \emph{Proceedings of the International Conference on Learning
  Representations (ICLR)}, 2018.

\bibitem[Bengio et~al.(2003)Bengio, Ducharme, Vincent, and
  Jauvin]{bengio2003neural}
Bengio, Y., Ducharme, R., Vincent, P., and Jauvin, C.
\newblock A neural probabilistic language model.
\newblock \emph{Journal of Machine Learning Research (JMLR)}, 2003.

\bibitem[Brown et~al.(2020)Brown, Mann, Ryder, Subbiah, Kaplan, Dhariwal,
  Neelakantan, Shyam, Sastry, Askell, et~al.]{brown2020language}
Brown, T.~B., Mann, B., Ryder, N., Subbiah, M., Kaplan, J., Dhariwal, P.,
  Neelakantan, A., Shyam, P., Sastry, G., Askell, A., et~al.
\newblock Language models are few-shot learners.
\newblock \emph{arXiv preprint arXiv:2005.14165}, 2020.

\bibitem[Chen et~al.(2020)Chen, Radford, Child, Wu, Jun, Luan, and
  Sutskever]{chen2020generative}
Chen, M., Radford, A., Child, R., Wu, J., Jun, H., Luan, D., and Sutskever, I.
\newblock Generative pretraining from pixels.
\newblock In \emph{Proceedings of the International Conference on Machine
  Learning (ICML)}, 2020.

\bibitem[Child et~al.(2019)Child, Gray, Radford, and
  Sutskever]{child2019generating}
Child, R., Gray, S., Radford, A., and Sutskever, I.
\newblock Generating long sequences with sparse {Transformers}.
\newblock \emph{arXiv preprint arXiv:1904.10509}, 2019.

\bibitem[Choromanski et~al.()Choromanski, Likhosherstov, Dohan, Song, Gane,
  Sarlos, Hawkins, Davis, Mohiuddin, Kaiser, Belanger, Colwell, and
  Weller]{choromanski2021rethinking}
Choromanski, K.~M., Likhosherstov, V., Dohan, D., Song, X., Gane, A., Sarlos,
  T., Hawkins, P., Davis, J.~Q., Mohiuddin, A., Kaiser, L., Belanger, D.~B.,
  Colwell, L.~J., and Weller, A.
\newblock Rethinking attention with {Performers}.
\newblock In \emph{Proceedings of the International Conference on Learning
  Representations (ICLR)}.

\bibitem[Clark et~al.(2021)Clark, Garrette, Turc, and Wieting]{clark2021canine}
Clark, J.~H., Garrette, D., Turc, I., and Wieting, J.
\newblock {CANINE:} pre-training an efficient tokenization-free encoder for
  language representation.
\newblock \emph{arXiv preprint arXiv:2103.06874}, 2021.

\bibitem[Dai et~al.(2019)Dai, Yang, Yang, Carbonell, Le, and
  Salakhutdinov]{dai2019transformerxl}
Dai, Z., Yang, Z., Yang, Y., Carbonell, J., Le, Q.~V., and Salakhutdinov, R.
\newblock {Transformer-XL}: Attentive language models beyond a fixed-length
  context.
\newblock In \emph{Proceedings of the Annual Meetings of the Association for
  Computational Linguistics (ACL)}, 2019.

\bibitem[Dai et~al.(2020)Dai, Lai, Yang, and Le]{dai2020funnel}
Dai, Z., Lai, G., Yang, Y., and Le, Q.
\newblock Funnel-{Transformer}: Filtering out sequential redundancy for
  efficient language processing.
\newblock In \emph{Proceedings of Neural Information Processing Systems
  (NeurIPS)}, 2020.

\bibitem[Dhariwal et~al.(2020)Dhariwal, Jun, Payne, Kim, Radford, and
  Sutskever]{dhariwal2020jukebox}
Dhariwal, P., Jun, H., Payne, C., Kim, J.~W., Radford, A., and Sutskever, I.
\newblock Jukebox: A generative model for music.
\newblock \emph{arXiv preprint arXiv:2005.00341}, 2020.

\bibitem[Dosovitskiy et~al.(2021)Dosovitskiy, Beyer, Kolesnikov, Weissenborn,
  Zhai, Unterthiner, Dehghani, Minderer, Heigold, Gelly,
  et~al.]{dosovitskiy2020image}
Dosovitskiy, A., Beyer, L., Kolesnikov, A., Weissenborn, D., Zhai, X.,
  Unterthiner, T., Dehghani, M., Minderer, M., Heigold, G., Gelly, S., et~al.
\newblock An image is worth 16x16 words: {Transformers} for image recognition
  at scale.
\newblock In \emph{Proceedings of the International Conference on Learning
  Representations (ICLR)}, 2021.

\bibitem[Graves(2013)]{graves2013generating}
Graves, A.
\newblock Generating sequences with recurrent neural networks.
\newblock \emph{arXiv preprint arXiv:1308.0850}, 2013.

\bibitem[Graves et~al.(2014)Graves, Wayne, and Danihelka]{graves2014neural}
Graves, A., Wayne, G., and Danihelka, I.
\newblock {Neural Turing Machines}.
\newblock \emph{arXiv preprint arXiv:1410.5401}, 2014.

\bibitem[Graves et~al.(2016)Graves, Wayne, Reynolds, Harley, Danihelka,
  Grabska-Barwi\'{n}ska, Colmenarejo, Grefenestette, Ramalho, Agapiou, Badia,
  Hermann, Zwols, Ostrovski, Cain, King, Summerfield, Blunsom, Kavukcuoglu, and
  Hassabis]{graves2016dnc}
Graves, A., Wayne, G., Reynolds, M., Harley, T., Danihelka, I.,
  Grabska-Barwi\'{n}ska, A., Colmenarejo, S.~G., Grefenestette, E., Ramalho,
  T., Agapiou, J., Badia, A.~P., Hermann, K.~M., Zwols, Y., Ostrovski, G.,
  Cain, A., King, H., Summerfield, C., Blunsom, P., Kavukcuoglu, K., and
  Hassabis, D.
\newblock Hybrid computing using a neural network with dynamic external memory.
\newblock \emph{Nature}, 538:\penalty0 471--476, 2016.

\bibitem[Gu et~al.(2021)Gu, Goel, and R\'{e}]{gu2021s4}
Gu, A., Goel, K., and R\'{e}, C.
\newblock Efficiently modeling long sequences with structured state spaces.
\newblock \emph{arXiv preprint arXiv:2111.00396}, 2021.

\bibitem[Harris et~al.(2020)Harris, Millman, van~der Walt, Gommers, Virtanen,
  Cournapeau, Wieser, Taylor, Berg, Smith, Kern, Picus, Hoyer, van Kerkwijk,
  Brett, Haldane, del R{\'{i}}o, Wiebe, Peterson, G{\'{e}}rard-Marchant,
  Sheppard, Reddy, Weckesser, Abbasi, Gohlke, and Oliphant]{harris2020array}
Harris, C.~R., Millman, K.~J., van~der Walt, S.~J., Gommers, R., Virtanen, P.,
  Cournapeau, D., Wieser, E., Taylor, J., Berg, S., Smith, N.~J., Kern, R.,
  Picus, M., Hoyer, S., van Kerkwijk, M.~H., Brett, M., Haldane, A., del
  R{\'{i}}o, J.~F., Wiebe, M., Peterson, P., G{\'{e}}rard-Marchant, P.,
  Sheppard, K., Reddy, T., Weckesser, W., Abbasi, H., Gohlke, C., and Oliphant,
  T.~E.
\newblock Array programming with {NumPy}.
\newblock \emph{Nature}, 585\penalty0 (7825):\penalty0 357--362, 2020.

\bibitem[Hawthorne et~al.(2018)Hawthorne, Elsen, Song, Roberts, Simon, Raffel,
  Engel, Oore, and Eck]{hawthorne2017onsets}
Hawthorne, C., Elsen, E., Song, J., Roberts, A., Simon, I., Raffel, C., Engel,
  J., Oore, S., and Eck, D.
\newblock Onsets and frames: Dual-objective piano transcription.
\newblock In \emph{Proceedings of the International Society for Music
  Information Retrieval Conference (ISMIR)}, 2018.

\bibitem[Hawthorne et~al.(2019)Hawthorne, Stasyuk, Roberts, Simon, Huang,
  Dieleman, Elsen, Engel, and Eck]{hawthorne2018enabling}
Hawthorne, C., Stasyuk, A., Roberts, A., Simon, I., Huang, C.-Z.~A., Dieleman,
  S., Elsen, E., Engel, J., and Eck, D.
\newblock Enabling factorized piano music modeling and generation with the
  {MAESTRO} dataset.
\newblock In \emph{Proceedings of the International Conference on Learning
  Representations (ICLR)}, 2019.

\bibitem[Hendrycks \& Gimpel(2016)Hendrycks and Gimpel]{hendrycks2016gelu}
Hendrycks, D. and Gimpel, K.
\newblock Gaussian error linear units {(GELUs)}.
\newblock \emph{arXiv preprint arXiv:1606.08415}, 2016.

\bibitem[Hessel et~al.(2020)Hessel, Budden, Viola, Rosca, Sezener, and
  Hennigan]{optax2020github}
Hessel, M., Budden, D., Viola, F., Rosca, M., Sezener, E., and Hennigan, T.
\newblock Optax: composable gradient transformation and optimisation, in
  {JAX}!, 2020.
\newblock URL \url{http://github.com/deepmind/optax}.

\bibitem[Huang et~al.(2019)Huang, Vaswani, Uszkoreit, Shazeer, Simon,
  Hawthorne, Dai, Hoffman, Dinculescu, and Eck]{huang2018music}
Huang, C.-Z.~A., Vaswani, A., Uszkoreit, J., Shazeer, N., Simon, I., Hawthorne,
  C., Dai, A.~M., Hoffman, M.~D., Dinculescu, M., and Eck, D.
\newblock {Music Transformer}: Generating music with long-term structure.
\newblock In \emph{Proceedings of the International Conference on Learning
  Representations (ICLR)}, 2019.

\bibitem[Jaegle et~al.(2021)Jaegle, Gimeno, Brock, Zisserman, Vinyals, and
  Carreira]{jaegle2021perceiver}
Jaegle, A., Gimeno, F., Brock, A., Zisserman, A., Vinyals, O., and Carreira, J.
\newblock Perceiver: General perception with iterative attention.
\newblock In \emph{Proceedings of the International Conference on Machine
  Learning (ICML)}, 2021.

\bibitem[Jaegle et~al.(2022)Jaegle, Borgeaud, Alayrac, Doersch, Ionescu, Ding,
  Koppula, Zoran, Brock, Shelhamer, Henaff, Botvinick, Zisserman, Vinyals, and
  Carreira]{jaegle2021io}
Jaegle, A., Borgeaud, S., Alayrac, J.-B., Doersch, C., Ionescu, C., Ding, D.,
  Koppula, S., Zoran, D., Brock, A., Shelhamer, E., Henaff, O., Botvinick,
  M.~M., Zisserman, A., Vinyals, O., and Carreira, J.
\newblock Perceiver {IO}: A general architecture for structured inputs \&
  outputs.
\newblock In \emph{Proceedings of the International Conference on Learning
  Representations (ICLR)}, 2022.

\bibitem[Jumper et~al.(2021)Jumper, Evans, Pritzel, Green, Figurnov,
  Ronneberger, Tunyasuvunakool, Bates, Zídek, Potapenko, Bridgland, Meyer,
  Kohl, Ballard, Cowie, Romera-Paredes, Nikolov, Jain, Adler, Back, Petersen,
  Reiman, Clancy, Zielinski, Steinegger, Pacholska, Berghammer, Bodenstein,
  Silver, Vinyals, Senior, Kavukcuoglu, Kohli, and Hassabis]{jumper2021highly}
Jumper, J., Evans, R., Pritzel, A., Green, T., Figurnov, M., Ronneberger, O.,
  Tunyasuvunakool, K., Bates, R., Zídek, A., Potapenko, A., Bridgland, A.,
  Meyer, C., Kohl, S. A.~A., Ballard, A.~J., Cowie, A., Romera-Paredes, B.,
  Nikolov, S., Jain, R., Adler, J., Back, T., Petersen, S., Reiman, D., Clancy,
  E., Zielinski, M., Steinegger, M., Pacholska, M., Berghammer, T., Bodenstein,
  S., Silver, D., Vinyals, O., Senior, A.~W., Kavukcuoglu, K., Kohli, P., and
  Hassabis, D.
\newblock Highly accurate protein structure prediction with {AlphaFold}.
\newblock \emph{Nature}, 596:\penalty0 583--589, 2021.

\bibitem[Katharopoulos et~al.(2020)Katharopoulos, Vyas, Pappas, and
  Fleuret]{katharopoulos2020transformers}
Katharopoulos, A., Vyas, A., Pappas, N., and Fleuret, F.
\newblock Transformers are {RNNs}: Fast autoregressive {Transformers} with
  linear attention.
\newblock In \emph{Proceedings of the International Conference on Machine
  Learning (ICML)}, 2020.

\bibitem[Kingma \& Ba(2015)Kingma and Ba]{kingma2015adam}
Kingma, D.~P. and Ba, J.
\newblock Adam: A method for stochastic optimization.
\newblock In \emph{Proceedings of the International Conference on Learning
  Representations (ICLR)}, 2015.

\bibitem[Kingma et~al.(2021)Kingma, Salimans, Poole, and Ho]{kingma2021vdm}
Kingma, D.~P., Salimans, T., Poole, B., and Ho, J.
\newblock On density estimation with diffusion models.
\newblock In \emph{Proceedings of Neural Information Processing Systems
  (NeurIPS)}, 2021.

\bibitem[Kitaev et~al.(2020)Kitaev, Kaiser, and Levskaya]{kitaev2020reformer}
Kitaev, N., Kaiser, L., and Levskaya, A.
\newblock Reformer: The efficient transformer.
\newblock In \emph{Proceedings of the International Conference on Learning
  Representations (ICLR)}, 2020.

\bibitem[Kudo \& Richardson(2018)Kudo and Richardson]{kudo2018sentencepiece}
Kudo, T. and Richardson, J.
\newblock {S}entence{P}iece: A simple and language independent subword
  tokenizer and detokenizer for neural text processing.
\newblock In \emph{Proceedings of the Annual Meetings of the Association for
  Computational Linguistics (ACL)}, 2018.

\bibitem[Lakhotia et~al.(2021)Lakhotia, Kharitonov, Hsu, Adi, Polyak, Bolte,
  Nguyen, Copet, Baevski, Mohamed, and Dupoux]{lakhotia2021generative}
Lakhotia, K., Kharitonov, E., Hsu, W.-N., Adi, Y., Polyak, A., Bolte, B.,
  Nguyen, T.-A., Copet, J., Baevski, A., Mohamed, A., and Dupoux, E.
\newblock Generative spoken language modeling from raw audio.
\newblock \emph{arXiv preprint arXiv:2102.01192}, 2021.

\bibitem[Lee et~al.(2019)Lee, Lee, Kim, Kosiorek, Choi, and Teh]{lee2019set}
Lee, J., Lee, Y., Kim, J., Kosiorek, A., Choi, S., and Teh, Y.~W.
\newblock Set {Transformer}: A framework for attention-based
  permutation-invariant neural networks.
\newblock In \emph{Proceedings of the International Conference on Machine
  Learning (ICML)}, 2019.

\bibitem[Liu et~al.(2018)Liu, Saleh, Pot, Goodrich, Sepassi, Kaiser, and
  Shazeer]{liu2018generating}
Liu, P.~J., Saleh, M., Pot, E., Goodrich, B., Sepassi, R., Kaiser, L., and
  Shazeer, N.
\newblock Generating {Wikipedia} by summarizing long sequences.
\newblock In \emph{Proceedings of the International Conference on Learning
  Representations (ICLR)}, 2018.

\bibitem[Ma et~al.(2021)Ma, Kong, Wang, Zhou, May, Ma, and
  Zettlemoyer]{ma2021luna}
Ma, X., Kong, X., Wang, S., Zhou, C., May, J., Ma, H., and Zettlemoyer, L.
\newblock {LUNA}: Linear unified nested attention.
\newblock In \emph{Proceedings of Neural Information Processing Systems
  (NeurIPS)}, 2021.

\bibitem[Mehri et~al.(2017)Mehri, Kumar, Gulrajani, Kumar, Jain, Sotelo,
  Courville, and Bengio]{mehri2017samplernn}
Mehri, S., Kumar, K., Gulrajani, I., Kumar, R., Jain, S., Sotelo, J.,
  Courville, A., and Bengio, Y.
\newblock {SampleRNN}: An unconditional end-to-end neural audio generation
  model.
\newblock In \emph{Proceedings of the International Conference on Learning
  Representations (ICLR)}, 2017.

\bibitem[Merity et~al.(2017)Merity, Xiong, Bradbury, and
  Socher]{merity2016pointer}
Merity, S., Xiong, C., Bradbury, J., and Socher, R.
\newblock Pointer sentinel mixture models.
\newblock In \emph{Proceedings of the International Conference on Learning
  Representations (ICLR)}, 2017.

\bibitem[Nash et~al.(2021)Nash, Menick, Dieleman, and
  Battaglia]{nash2021generating}
Nash, C., Menick, J., Dieleman, S., and Battaglia, P.~W.
\newblock Generating images with sparse representations.
\newblock In \emph{Proceedings of the International Conference on Machine
  Learning (ICML)}, 2021.

\bibitem[Nawrot et~al.(2021)Nawrot, Tworkowski, Tyrolski, Kaiser, Wu, Szegedy,
  and Michalewski]{nawrot2021hierarchical}
Nawrot, P., Tworkowski, S., Tyrolski, M., Kaiser, L., Wu, Y., Szegedy, C., and
  Michalewski, H.
\newblock Hierarchical {Transformers} are more efficient language models.
\newblock \emph{arXiv preprint arXiv:2110.13711}, 2021.

\bibitem[Peng et~al.(2021)Peng, Pappas, Yogatama, Schwartz, Smith, and
  Kong]{peng2021random}
Peng, H., Pappas, N., Yogatama, D., Schwartz, R., Smith, N., and Kong, L.
\newblock Random feature attention.
\newblock In \emph{Proceedings of the International Conference on Learning
  Representations (ICLR)}, 2021.

\bibitem[Polyak et~al.(2021)Polyak, Adi, Copet, Kharitonov, Lakhotia, Hsu,
  Mohamed, and Dupoux]{polyak2021speech}
Polyak, A., Adi, Y., Copet, J., Kharitonov, E., Lakhotia, K., Hsu, W.-N.,
  Mohamed, A., and Dupoux, E.
\newblock Speech resynthesis from discrete disentangled self-supervised
  representations.
\newblock \emph{arXiv preprint arXiv:2104.00355}, 2021.

\bibitem[Press et~al.(2021)Press, Smith, and Lewis]{press2021shortformer}
Press, O., Smith, N.~A., and Lewis, M.
\newblock Shortformer: Better language modeling using shorter inputs.
\newblock In \emph{Proceedings of the Annual Meetings of the Association for
  Computational Linguistics (ACL)}, 2021.

\bibitem[Rabe \& Staats(2021)Rabe and Staats]{rabe2021self}
Rabe, M.~N. and Staats, C.
\newblock Self-attention does not need {$O(n^2)$} memory.
\newblock \emph{arXiv preprint arXiv:2112.05682}, 2021.

\bibitem[Rae et~al.(2019)Rae, Potapenko, Jayakumar, Hillier, and
  Lillicrap]{rae2019compressive}
Rae, J.~W., Potapenko, A., Jayakumar, S.~M., Hillier, C., and Lillicrap, T.~P.
\newblock Compressive {Transformers} for long-range sequence modelling.
\newblock In \emph{Proceedings of the International Conference on Learning
  Representations (ICLR)}, 2019.

\bibitem[Rae et~al.(2021)Rae, Borgeaud, Cai, Millican, Hoffmann, Song,
  Aslanides, Henderson, Ring, Young, Rutherford, Hennigan, Menick, Cassirer,
  Powell, van~den Driessche, Hendricks, Rauh, Huang, Glaese, Welbl, Dathathri,
  Huang, Uesato, Mellor, Higgins, Creswell, McAleese, Wu, Elsen, Jayakumar,
  Buchatskaya, Budden, Sutherland, Simonyan, Paganini, Sifre, Martens, Li,
  Kuncoro, Nematzadeh, Gribovskaya, Donato, Lazaridou, Mensch, Lespiau,
  Tsimpoukelli, Grigorev, Fritz, Sottiaux, Pajarskas, Pohlen, Gong, Toyama,
  de~Masson~d’Autume, Li, Terzi, Mikulik, Babuschkin, Clark, de~Las~Casas,
  Guy, Jones, Bradbury, Johnson, Hechtman, Weidinger, Gabriel, Isaac, Lockhart,
  Osindero, Rimell, Dyer, Vinyals, Ayoub, Stanway, Bennett, Hassabis,
  Kavukcuoglu, and Irving]{rae2021scaling}
Rae, J.~W., Borgeaud, S., Cai, T., Millican, K., Hoffmann, J., Song, F.,
  Aslanides, J., Henderson, S., Ring, R., Young, S., Rutherford, E., Hennigan,
  T., Menick, J., Cassirer, A., Powell, R., van~den Driessche, G., Hendricks,
  L.~A., Rauh, M., Huang, P.-S., Glaese, A., Welbl, J., Dathathri, S., Huang,
  S., Uesato, J., Mellor, J., Higgins, I., Creswell, A., McAleese, N., Wu, A.,
  Elsen, E., Jayakumar, S., Buchatskaya, E., Budden, D., Sutherland, E.,
  Simonyan, K., Paganini, M., Sifre, L., Martens, L., Li, X.~L., Kuncoro, A.,
  Nematzadeh, A., Gribovskaya, E., Donato, D., Lazaridou, A., Mensch, A.,
  Lespiau, J.-B., Tsimpoukelli, M., Grigorev, N., Fritz, D., Sottiaux, T.,
  Pajarskas, M., Pohlen, T., Gong, Z., Toyama, D., de~Masson~d’Autume, C.,
  Li, Y., Terzi, T., Mikulik, V., Babuschkin, I., Clark, A., de~Las~Casas, D.,
  Guy, A., Jones, C., Bradbury, J., Johnson, M., Hechtman, B., Weidinger, L.,
  Gabriel, I., Isaac, W., Lockhart, E., Osindero, S., Rimell, L., Dyer, C.,
  Vinyals, O., Ayoub, K., Stanway, J., Bennett, L., Hassabis, D., Kavukcuoglu,
  K., and Irving, G.
\newblock Scaling language models: Methods, analysis \& insights from training
  {Gopher}.
\newblock \emph{arXiv preprint arXiv:2112.11446}, 2021.

\bibitem[Raffel et~al.(2020)Raffel, Shazeer, Roberts, Lee, Narang, Matena,
  Zhou, Li, and Liu]{raffell2020exploring}
Raffel, C., Shazeer, N., Roberts, A., Lee, K., Narang, S., Matena, M., Zhou,
  Y., Li, W., and Liu, P.~J.
\newblock Exploring the limits of transfer learning with a unified text-to-text
  {Transformer}.
\newblock \emph{Journal of Machine Learning Research (JMLR)}, 2020.

\bibitem[Rajbhandari et~al.(2020)Rajbhandari, Rasley, Ruwase, and
  He]{Rajbhandari2020ZeRO}
Rajbhandari, S., Rasley, J., Ruwase, O., and He, Y.
\newblock {ZeRO}: Memory optimizations toward training trillion parameter
  models.
\newblock In \emph{Proceedings of the International Conference for High
  Performance Computing, Networking, Storage and Analysis (SC)}, 2020.

\bibitem[Ramesh et~al.(2021)Ramesh, Pavlov, Goh, Gray, Voss, Radford, Chen, and
  Sutskever]{ramesh2021zero}
Ramesh, A., Pavlov, M., Goh, G., Gray, S., Voss, C., Radford, A., Chen, M., and
  Sutskever, I.
\newblock Zero-shot text-to-image generation.
\newblock \emph{arXiv preprint arXiv:2102.12092}, 2021.

\bibitem[Ren et~al.(2021)Ren, Dai, Dai, Yang, Leskovec, Schuurmans, and
  Dai]{ren2021combiner}
Ren, H., Dai, H., Dai, Z., Yang, M., Leskovec, J., Schuurmans, D., and Dai, B.
\newblock Combiner: Full attention {Transformer} with sparse computation cost.
\newblock In \emph{Proceedings of Neural Information Processing Systems
  (NeurIPS)}, 2021.

\bibitem[Rosenfeld(2000)]{rosenfeld2000two}
Rosenfeld, R.
\newblock Two decades of statistical language modeling: where do we go from
  here?
\newblock \emph{Proceedings of the IEEE}, 88\penalty0 (8):\penalty0 1270--1278,
  2000.

\bibitem[Roy et~al.(2021)Roy, Saffar, Vaswani, and Grangier]{roy2021routing}
Roy, A., Saffar, M., Vaswani, A., and Grangier, D.
\newblock Efficient content-based sparse attention with {Routing Transformers}.
\newblock \emph{Transactions of the Association for Computational Linguistics
  (TACL)}, 9, 2021.

\bibitem[Saharia et~al.(2021)Saharia, Ho, Chan, Salimans, Fleet, and
  Norouzi]{saharia2021image}
Saharia, C., Ho, J., Chan, W., Salimans, T., Fleet, D.~J., and Norouzi, M.
\newblock Image super-resolution via iterative refinement.
\newblock \emph{arXiv preprint arXiv:2104.07636}, 2021.

\bibitem[Schmidhuber \& Heil(1994)Schmidhuber and
  Heil]{schmidhuber1994sequential}
Schmidhuber, J. and Heil, S.
\newblock Sequential neural text compression.
\newblock \emph{IEEE Transactions on Neural Networks}, 7\penalty0 (1):\penalty0
  142--146, 1994.

\bibitem[Shazeer(2019)]{shazeer2019fast}
Shazeer, N.
\newblock Fast {Transformer} decoding: one write-head is all you need.
\newblock \emph{arXiv preprint arXiv:1911.02150}, 2019.

\bibitem[Simon et~al.(2019)Simon, Huang, Engel, Hawthorne, and
  Dinculescu]{simon2019}
Simon, I., Huang, C.-Z.~A., Engel, J., Hawthorne, C., and Dinculescu, M.
\newblock Generating piano music with transformer.
\newblock 2019.
\newblock URL \url{https://magenta.tensorflow.org/piano-transformer}.

\bibitem[So et~al.(2021)So, Mańke, Liu, Dai, Shazeer, and Le]{so2021primer}
So, D.~R., Mańke, W., Liu, H., Dai, Z., Shazeer, N., and Le, Q.~V.
\newblock Primer: Searching for efficient {Transformers} for language modeling.
\newblock \emph{arXiv preprint arXiv:2109.08668}, 2021.

\bibitem[Srivastava et~al.(2014)Srivastava, Hinton, Krizhevsky, Sutskever, and
  Salakhutdinov]{JMLR:v15:srivastava14a}
Srivastava, N., Hinton, G., Krizhevsky, A., Sutskever, I., and Salakhutdinov,
  R.
\newblock Dropout: A simple way to prevent neural networks from overfitting.
\newblock \emph{Journal of Machine Learning Research (JMLR)}, 2014.

\bibitem[Su et~al.(2021)Su, Lu, Pan, Wen, and Liu]{su2021roformer}
Su, J., Lu, Y., Pan, S., Wen, B., and Liu, Y.
\newblock {RoFormer}: Enhanced {Transformer} with rotary position embedding.
\newblock \emph{arXiv preprint arxiv:2104.09864}, 2021.

\bibitem[Sun et~al.(2021)Sun, Krishna, Mattarella-Micke, and
  Iyyer]{sun2021long_range}
Sun, S., Krishna, K., Mattarella-Micke, A., and Iyyer, M.
\newblock Do long-range language models actually use long-range context?
\newblock In \emph{Proceedings of the Annual Conference on Empirical Methods in
  Natural Language Processing (EMNLP)}, 2021.

\bibitem[Sutskever et~al.(2014)Sutskever, Vinyals, and
  Le]{sutskever2014seq2seq}
Sutskever, I., Vinyals, O., and Le, Q.~V.
\newblock Sequence to sequence learning with neural networks.
\newblock In \emph{Proceedings of Neural Information Processing Systems
  (NeurIPS)}, 2014.

\bibitem[Uria et~al.(2016)Uria, C\^{o}t\'{e}, Gregor, Murray, and
  Larochelle]{uria2016neural}
Uria, B., C\^{o}t\'{e}, M.-A., Gregor, K., Murray, I., and Larochelle, H.
\newblock Neural autoregressive distribution estimation.
\newblock \emph{Journal of Machine Learning Research (JMLR)}, 2016.

\bibitem[{van den Oord} et~al.(2016{\natexlab{a}}){van den Oord}, Dieleman,
  Zen, Simonyan, Vinyals, Graves, Kalchbrenner, Senior, and
  Kavukcuoglu]{oord2016wavenet}
{van den Oord}, A., Dieleman, S., Zen, H., Simonyan, K., Vinyals, O., Graves,
  A., Kalchbrenner, N., Senior, A., and Kavukcuoglu, K.
\newblock {WaveNet}: A generative model for raw audio.
\newblock \emph{arXiv preprint arXiv:1609.03499}, 2016{\natexlab{a}}.

\bibitem[{van den Oord} et~al.(2016{\natexlab{b}}){van den Oord}, Kalchbrenner,
  and Kavukcuoglu]{oord2016pixel}
{van den Oord}, A., Kalchbrenner, N., and Kavukcuoglu, K.
\newblock Pixel recurrent neural networks.
\newblock In \emph{Proceedings of the International Conference on Machine
  Learning (ICML)}, 2016{\natexlab{b}}.

\bibitem[van~den Oord et~al.(2017)van~den Oord, Vinyals, and
  Kavukcuoglu]{oord17discrete}
van~den Oord, A., Vinyals, O., and Kavukcuoglu, K.
\newblock Neural discrete representation learning.
\newblock In \emph{Proceedings of Neural Information Processing Systems
  (NeurIPS)}, 2017.

\bibitem[Vaswani et~al.()Vaswani, Bengio, Brevdo, Chollet, Gomez, Gouws, Jones,
  Kaiser, Kalchbrenner, Parmar, Sepassi, Shazeer, and Uszkoreit]{tensor2tensor}
Vaswani, A., Bengio, S., Brevdo, E., Chollet, F., Gomez, A.~N., Gouws, S.,
  Jones, L., Kaiser, L., Kalchbrenner, N., Parmar, N., Sepassi, R., Shazeer,
  N., and Uszkoreit, J.
\newblock Tensor2tensor for neural machine translation.
\newblock \emph{arXiv preprint arXiv:1803.07416}.

\bibitem[Vaswani et~al.(2017)Vaswani, Shazeer, Parmar, Uszkoreit, Jones, Gomez,
  Kaiser, and Polosukhin]{vaswani2017attention}
Vaswani, A., Shazeer, N., Parmar, N., Uszkoreit, J., Jones, L., Gomez, A.~N.,
  Kaiser, L., and Polosukhin, I.
\newblock Attention is all you need.
\newblock In \emph{Proceedings of Neural Information Processing Systems
  (NeurIPS)}, 2017.

\bibitem[Vinyals et~al.(2019)Vinyals, Babuschkin, Czarnecki, Mathieu, Dudzik,
  Chung, Choi, Powell, Ewalds, Georgiev, Oh, Horgan, Kroiss, Danihelka, Huang,
  Sifre, Cai, Agapiou, Jaderberg, Vezhnevets, Leblond, Pohlen, Dalibard,
  Budden, Sulsky, Molloy, Paine, Gulcehre, Wang, Pfaff, Wu, Ring, Yogatama,
  McKinney, Smith, Schaul, Lillicrap, Kavukcuoglu, Hassabis, Chris, and
  Silver]{vinyals2019grandmaster}
Vinyals, O., Babuschkin, I., Czarnecki, W.~M., Mathieu, M., Dudzik, A., Chung,
  J., Choi, D.~H., Powell, R., Ewalds, T., Georgiev, P., Oh, J., Horgan, D.,
  Kroiss, M., Danihelka, I., Huang, A., Sifre, L., Cai, T., Agapiou, J.~P.,
  Jaderberg, M., Vezhnevets, A.~S., Leblond, R., Pohlen, T., Dalibard, V.,
  Budden, D., Sulsky, Y., Molloy, J., Paine, T.~L., Gulcehre, C., Wang, Z.,
  Pfaff, T., Wu, Y., Ring, R., Yogatama, Dani~W\"{u}nsch, D., McKinney, K.,
  Smith, O., Schaul, T., Lillicrap, T., Kavukcuoglu, K., Hassabis, D., Chris,
  A., and Silver, D.
\newblock Grandmaster level in {StarCraft} {II} using multi-agent reinforcement
  learning.
\newblock \emph{Nature}, 575\penalty0 (7782):\penalty0 350--354, 2019.

\bibitem[Wang(2021)]{kingoflolz2021github}
Wang, B.
\newblock Mesh transformer {Jax}, 2021.
\newblock URL \url{https://github.com/kingoflolz/mesh-transformer-jax}.

\bibitem[Wang et~al.(2020)Wang, Li, Khabsa, Fang, and Ma]{wang2020linformer}
Wang, S., Li, B.~Z., Khabsa, M., Fang, H., and Ma, H.
\newblock Linformer: Self-attention with linear complexity.
\newblock \emph{arXiv preprint arXiv:2006.04768}, 2020.

\bibitem[Weston et~al.(2015)Weston, Chopra, and Bordes]{weston2014memory}
Weston, J., Chopra, S., and Bordes, A.
\newblock Memory networks.
\newblock In \emph{Proceedings of the International Conference on Learning
  Representations (ICLR)}, 2015.

\bibitem[Wu et~al.(2021)Wu, Liang, Ji, Yang, Fang, Jiang, and Duan]{wu2021nuwa}
Wu, C., Liang, J., Ji, L., Yang, F., Fang, Y., Jiang, D., and Duan, N.
\newblock {N\"UWA}: Visual synthesis pre-training for neural visual world
  creation.
\newblock \emph{arXiv preprint arXiv:2111.12417}, 2021.

\bibitem[Wu et~al.(2019)Wu, Fan, Baevski, Dauphin, and Auli]{wu2019pay}
Wu, F., Fan, A., Baevski, A., Dauphin, Y.~N., and Auli, M.
\newblock Pay less attention with lightweight and dynamic convolutions.
\newblock \emph{arXiv preprint arXiv:1901.10430}, 2019.

\bibitem[Wu et~al.(2022)Wu, Rabe, Hutchins, and Szegedy]{wu2022memorizing}
Wu, Y., Rabe, M.~N., Hutchins, D., and Szegedy, C.
\newblock Memorizing transformers.
\newblock In \emph{Proceedings of the International Conference on Learning
  Representations (ICLR)}, 2022.
\newblock URL \url{https://openreview.net/forum?id=TrjbxzRcnf-}.

\bibitem[Xiong et~al.(2020)Xiong, Yang, He, Zheng, Zheng, Xing, Zhang, Lan,
  Wang, and Liu]{xiong2020layernorm}
Xiong, R., Yang, Y., He, D., Zheng, K., Zheng, S., Xing, C., Zhang, H., Lan,
  Y., Wang, L., and Liu, T.-Y.
\newblock On layer normalization in the {Transformer} architecture.
\newblock In \emph{Proceedings of the International Conference on Machine
  Learning (ICML)}, 2020.

\bibitem[Zaheer et~al.(2020)Zaheer, Guruganesh, Dubey, Ainslie, Alberti,
  Ontanon, Pham, Ravula, Wang, Yang, et~al.]{zaheer2020bigbird}
Zaheer, M., Guruganesh, G., Dubey, K.~A., Ainslie, J., Alberti, C., Ontanon,
  S., Pham, P., Ravula, A., Wang, Q., Yang, L., et~al.
\newblock {Big Bird: Transformers} for longer sequences.
\newblock \emph{Proceedings of Neural Information Processing Systems
  (NeurIPS)}, 33, 2020.

\bibitem[Zeghidour et~al.(2021)Zeghidour, Luebs, Omran, Skoglund, and
  Tagliasacchi]{zeghidour2021soundstream}
Zeghidour, N., Luebs, A., Omran, A., Skoglund, J., and Tagliasacchi, M.
\newblock {SoundStream}: An end-to-end neural audio codec.
\newblock \emph{IEEE/ACM Transactions on Audio, Speech, and Language Processing
  (TASLP)}, 2021.

\end{thebibliography}
